\documentclass[letterpaper, 10 pt, conference]{ieeeconf}  

\IEEEoverridecommandlockouts                              

\usepackage{url}
\usepackage{cite}
\usepackage{balance}
\usepackage{graphicx}
\usepackage{amsfonts}
\usepackage{fancyhdr}
\usepackage{comment}
\usepackage{times}
\usepackage{amsmath}
\usepackage{changepage}
\usepackage{amssymb}
\usepackage{enumerate} 
\usepackage{algorithmic}
\usepackage{bm}
\usepackage{calligra}
\usepackage{multirow}
\usepackage{tabularx}

\usepackage{booktabs}
\usepackage{mathtools}
\usepackage{arydshln}
\usepackage{latexsym} 
\usepackage{amsmath}
\usepackage{amssymb}
\usepackage{color}
\usepackage[ruled,vlined]{algorithm2e}
\usepackage{ulem}
\usepackage{float}
\usepackage{tabstackengine}
\usepackage{siunitx}
\usepackage{colortbl}
\usepackage{hhline}
\usepackage{bookmark}
\usepackage{dirtree}
\usepackage{subfigure}

\newcommand{\rt}{\textcolor[rgb]{1,0,0}}

\newcommand{\bt}{\textcolor[rgb]{0,0,1}}

\begin{document}

\title{\LARGE \bf
FusionPortable: A Multi-Sensor Campus-Scene Dataset for Evaluation of Localization and Mapping Accuracy on Diverse Platforms
}

\author{
Jianhao Jiao$^{2,4*}$, Hexiang Wei$^{2*}$, Tianshuai Hu$^{2*}$, Xiangcheng Hu$^{2*}$, Yilong Zhu$^{2}$, 
Zhijian He$^{2}$, Jin Wu$^{2}$, \\
Jingwen Yu$^{2,5}$, Xupeng Xie$^{2}$, 
Huaiyang Huang$^{2}$, Ruoyu Geng$^{2}$,
Lujia Wang$^{2,4}$, Ming Liu$^{1,2,3}$
\thanks{$^{*}$Equal contribution.}
\thanks{$^{1}$The Hong Kong University of Science and Technology (Guangzhou), Nansha, Guangzhou, 511400, Guangdong, China.}
\thanks{$^{2}$The Hong Kong University of Science and Technology, Hong Kong, China. {\tt\small jjiao@connect.ust.hk, eelium@ust.hk}.}
\thanks{$^{3}$HKUST Shenzhen-Hong Kong Collaborative Innovation Research Institute, Futian, Shenzhen, China.}
\thanks{$^{4}$Clear Water Bay Institute of Autonomous Driving, Hong Kong, China.}
\thanks{$^{5}$Department of Electronic and Electrical Engineering, Southern University of Science and Technology, Shenzhen, China.}
\thanks{This work was supported by Zhongshan Science and Technology Bureau Fund, under project 2020AG002, Foshan-HKUST Project no. FSUST20-SHCIRI06C, and the Project of Hetao Shenzhen-Hong Kong Science and Technology Innovation Cooperation Zone(HZQB-KCZYB-2020083), awarded to Prof. Ming Liu.}
}

\maketitle

\begin{abstract}
  Combining multiple sensors enables a robot to maximize its perceptual awareness of environments and enhance its robustness to external disturbance, crucial to robotic navigation.
  This paper proposes the FusionPortable benchmark, a complete multi-sensor dataset with a diverse set of sequences for mobile robots. This paper presents three contributions.
  We first advance a portable and versatile multi-sensor suite that offers rich sensory measurements: 10Hz LiDAR point clouds, 20Hz stereo frame images, high-rate and asynchronous events from stereo event cameras, 200Hz inertial readings from an IMU, and 10Hz GPS signal. Sensors are already temporally synchronized in hardware. 
  This device is lightweight, self-contained, and has plug-and-play support for mobile robots. 
  Second, we construct a dataset by collecting 17 sequences that cover a variety of environments on the campus by exploiting multiple robot platforms for data collection.
  Some sequences are challenging to existing SLAM algorithms.
  Third, we provide ground truth for the decouple localization and mapping performance evaluation. 
  We additionally evaluate state-of-the-art SLAM approaches and identify their limitations. 
  The dataset, consisting of raw sensor measurements, ground truth, calibration data, and evaluated algorithms, will be released.
\end{abstract}

\section{Introduction}
\label{sec:introduction}





\subsection{Motivation}
Multi-sensor fusion for robust perception is fundamental to various robotic applications.
Different sensors can complement each other, and thus the system's perception capability is enhanced with sensor fusion.
Over the past decades, research on multi-sensor SLAM has made substantial progress. 
High-quality open datasets, which are collections of multi-sensor data and provide a suite of benchmark tools, significantly contribute to this advancement.
On one hand, these datasets can waive inhibitive requirements on budget and workforce, such as system integration calibration and field operations. 
On the other hand, they investigate the advantages and limitations of current SLAM solutions and elaborately design practical, but challenging sequences \cite{pomerleau2012challenging, wang2020tartanair}. 
Several of them also introduce novel sensors and indicate future research opportunities \cite{mueggler2017event}. 
Researchers can easily develop, validate, and rank their algorithms with others, thus accelerating the breakthroughs.
However, existing datasets were mostly collected with a single data collection platform or simplified sensor configuration.
Researchers may only utilize limited sensors to develop algorithms that has a risk of over-fitting to a benchmark.
Hence, we consider that a desirable dataset should fulfill the following four requirements.
\begin{enumerate}
  \item Various sensors are required, making it possible to explore novel approaches to utilize them jointly. 
  \item Algorithm evaluation should be fairely conducted on various mobile robots. These robots perform different motion patterns that may challenge several SLAM algorithms' assumptions.
  \item Sequences have to cover from room-scale (meter-level) to large-scale (kilometer-level) environments to evaluate algorithms' scalability. 
  \item Ground-truth trajectories and 3D maps are required to evaluate algorithms' localization and surface reconstruction accuracy, respectively. 
\end{enumerate}

\begin{table*}[t]
	\centering
	\caption{Comparison with previous datasets on data-acquisition platform, environment, sensor type, and ground-truth method.}
	\vspace{-0.2cm}
	\renewcommand\arraystretch{1.0}
	\renewcommand\tabcolsep{2.7pt}
	\scriptsize
	\begin{tabular}{lllccccc|c|c}
	\toprule[0.03cm]
	\multirow{2}{*}{Dataset} & \multirow{2}{*}{Platform} & \multirow{2}{*}{Environment} & \multicolumn{5}{c}{Sensor} & \multirow{2}{*}{GT Pose} & \multirow{2}{*}{GT Map}\\ \cline{4-8}
	& & & IMU & GPS & LiDAR & Frame Cam. & Event Cam. &  \\
	\toprule[0.03cm]
	
  UZH-Event \cite{mueggler2017event} & Handheld & In/Outdoors & \checkmark & & & & \checkmark & Mocap & \\


	
	ETH-EuRoc \cite{burri2016euroc} & MAV & Indoors & \checkmark & & & \checkmark & & Mocap/LT & Nova MS50 \\

	TUM VI \cite{schubert2018tum} & Handheld & In/Outdoors & \checkmark & & & \checkmark & & Mocap &  \\


	\midrule[0.03cm]
	MIT DARPA \cite{huang2010high} & Car & Urban & \checkmark & \checkmark & \checkmark & \checkmark & & GPS/INS & \\

	KITTI \cite{geiger2013vision} & Car & Urban & \checkmark & \checkmark & \checkmark & \checkmark & & RTK-GPS/INS &  \\

	Oxford RobotCar \cite{maddern20171} & Car & Urban & \checkmark & \checkmark & \checkmark & \checkmark & & GPS/INS/SLAM & \\

	UrbanLoc \cite{wen2020urbanloco} & Car & Urban & \checkmark & \checkmark & \checkmark & \checkmark & & GPS/INS & \\

	\midrule[0.03cm]
	Newer College \cite{ramezani2020newer} & Handheld & Outdoors & \checkmark & & \checkmark & \checkmark & & 6DoF ICP & BLK$360$ \\

	NCLT \cite{carlevaris2016university} & UGV & In/Outdoors & \checkmark & \checkmark & \checkmark & \checkmark & & RTK-GPS/SLAM &  \\

	M2DGR \cite{yin2021m2dgr} & UGV & In/Outdoors & \checkmark & \checkmark & \checkmark & \checkmark & \checkmark & RTK-GPS/Mocap/LT & \\

	MVSEC \cite{zhu2018multivehicle} & Handheld/UAV/Motorcycle/Car & In/Outdoors & \checkmark & \checkmark & \checkmark & \checkmark & \checkmark & Mocap/SLAM & \\

	\midrule[0.03cm]
	\textbf{Ours (FusionPortable)} & Handheld/Quad. Robot/UGV & In/Outdoors & \checkmark & \checkmark & \checkmark & \checkmark & \checkmark & Mocap/RTK-GPS/6DoF NDT & BLK$360$ \\

  \bottomrule[0.03cm]

  \multicolumn{10}{l}{
    Mocap: Motion capture system. 
		LT: Laser tracker.
  }	
	\end{tabular}
	\label{tab:related_work_dataset}
	\vspace{-0.5cm}
\end{table*}

\subsection{Contributions}
There appears to be an absence of compatible public datasets that satisfy these requirements, motivating us to propose a new SLAM benchmark.

\begin{figure*}[t]
	\centering
	\subfigure[]
	{\label{fig:sensor_handheld}\centering\includegraphics[width=.2746\linewidth]{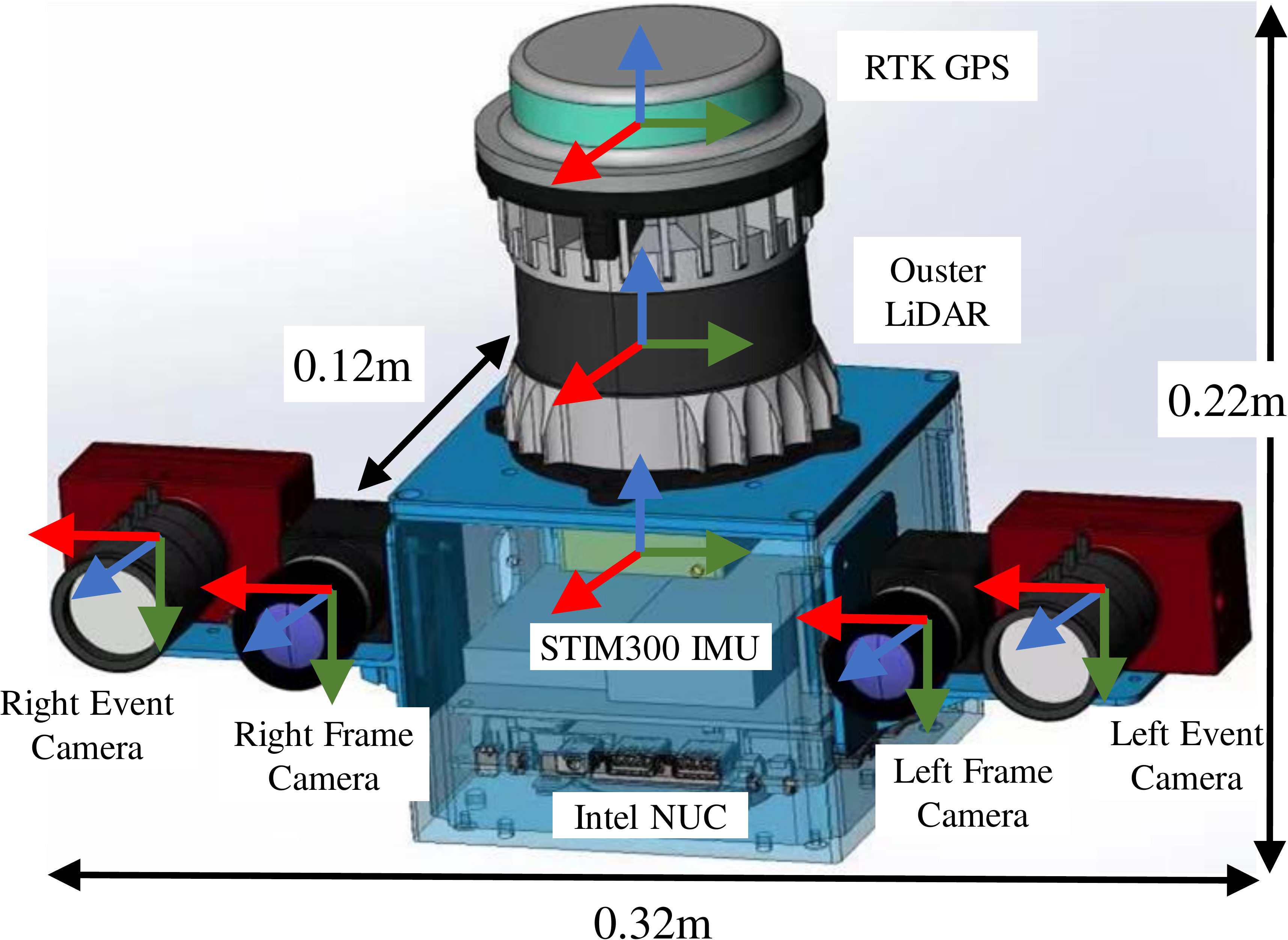}}
	\hspace{0.2cm}
	\subfigure[]
	{\label{fig:sensor_stab}\centering\includegraphics[width=.1701\linewidth]{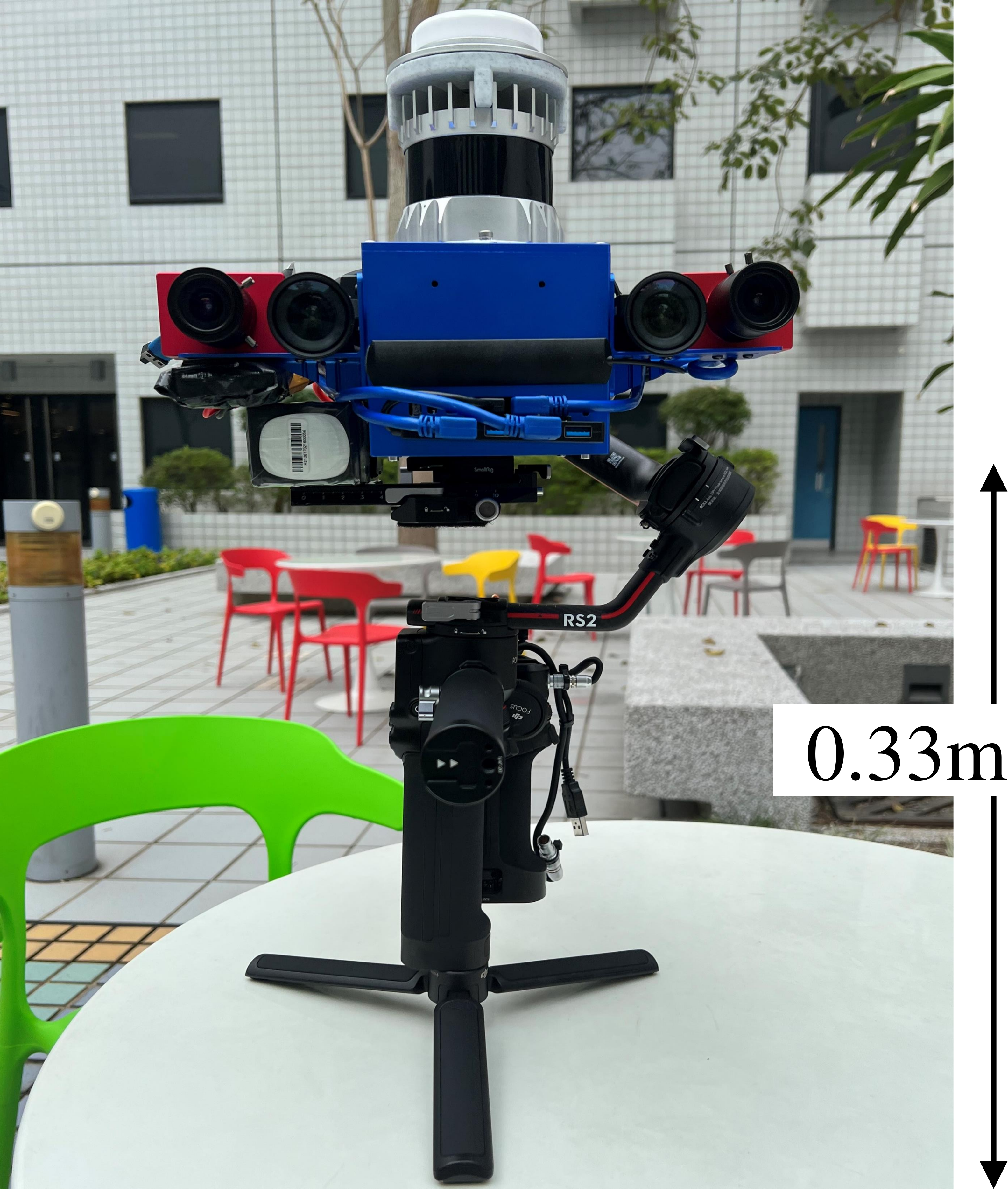}}
	\hspace{0.2cm}
	\subfigure[]
	{\label{fig:sensor_quadrobot}\centering\includegraphics[width=.1776\linewidth]{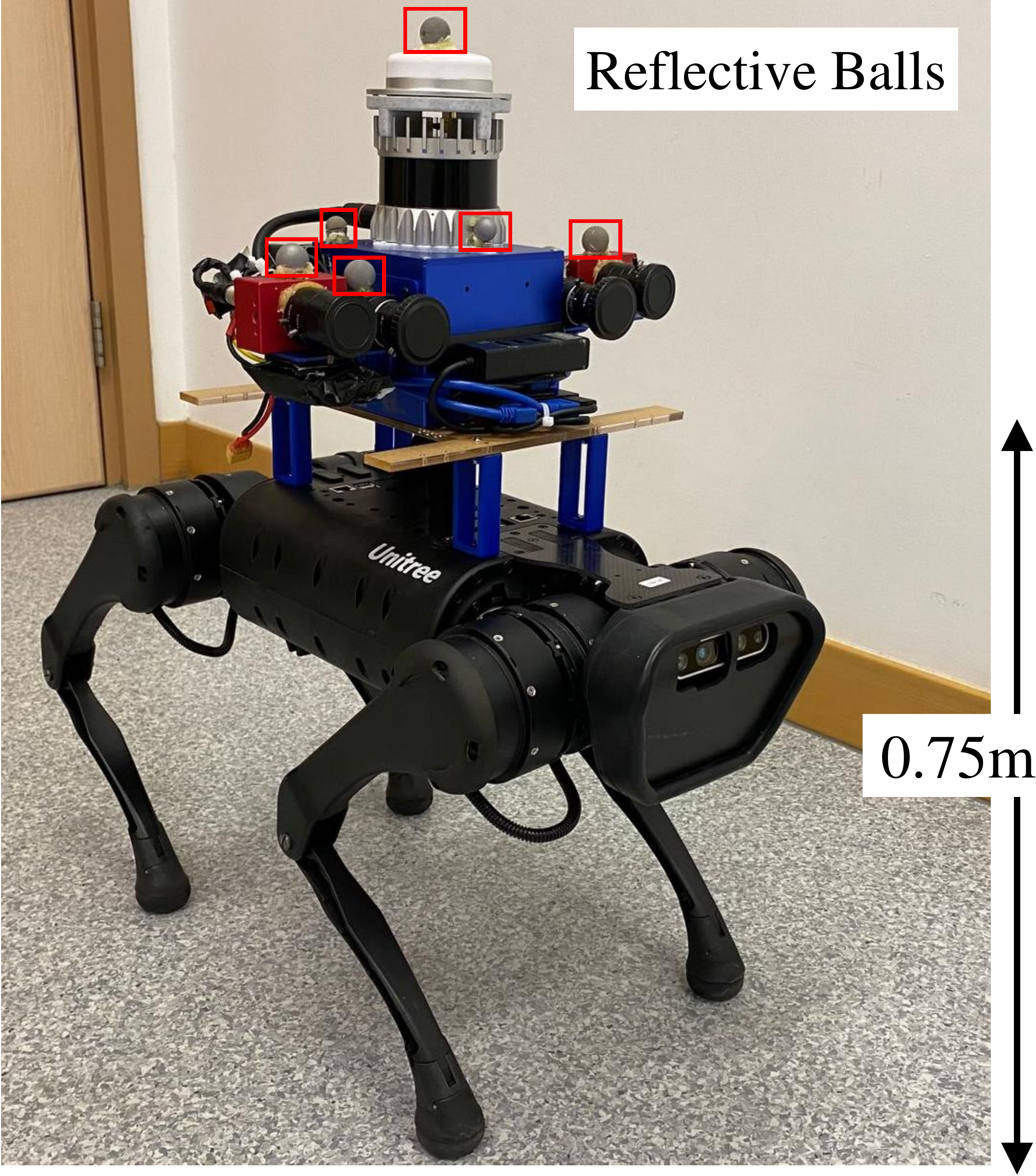}}
	\hspace{0.2cm}
	\subfigure[]
	{\label{fig:sensor_apollo}\centering\includegraphics[width=.1701\linewidth]{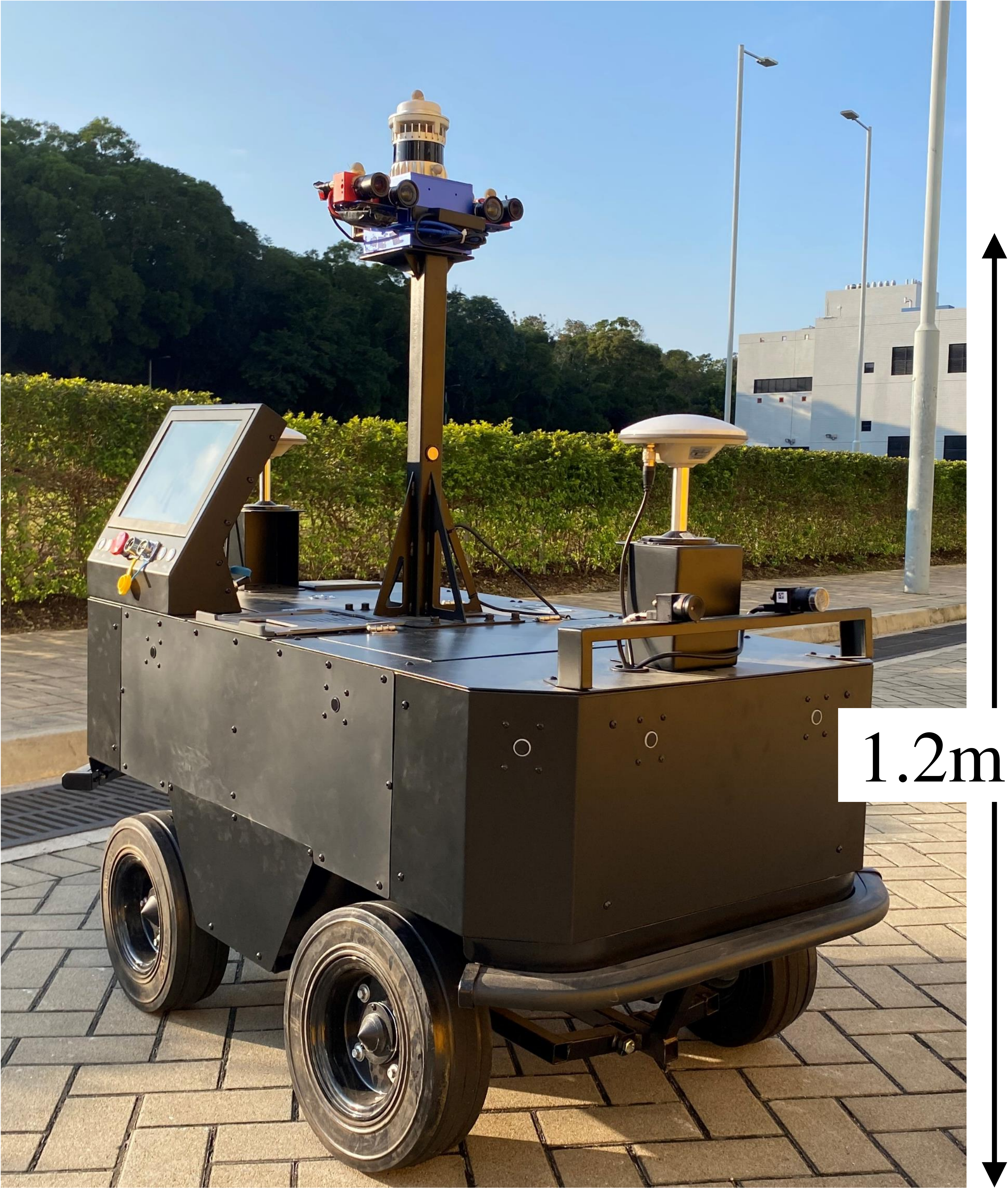}}
	\caption{The multi-sensor device and data collection platform: 
	(a) CAD model of the sensor rig, where axis directions are colored: red: $X$, green: $Y$, blue: $Z$. The sensor rig is rigidly mounted on (b) a gimbal stabilizer, (c) a quadruped robot, and (d) an apollo autonomous vehicle.}
	\label{fig:sensor_picture}
	\vspace{-0.3cm}
\end{figure*}

This paper proposes the \textbf{FusionPortable benchmark}, a novel multi-sensor dataset 
with a set of sequences from diverse environments.
Our contributions are presented three-fold.
First, a \textit{portable} and \textit{versatile} multi-sensor device is elaborately manufactured.
Two RGB frame cameras are mounted on the left and right side, one high-frequency and high-precision IMU is mounted internally, and one RTK-GPS is installed on the top position.
Moreover, thanks to current progress in sensory technology, both novel event cameras and high-resolution 3D LiDAR are available.
Thus, we also integrate them with our sensor rig and investigate their performance.
All these sensors are mounted on the same rigid aluminum-alloy-based parts. 
Thus, their spatial relation has a tiny dynamic deviation.
The complete device has its own clock synchronization unit, processor, and battery, thus self-contained.
Since its size, weight, and extensibility (see Fig. \ref{fig:sensor_picture}) are satisfying, we advance that it would be a plug-and-play support to various mobile robots.

Second, we install the sensor rig on various platforms ranging from the handheld mode with a gimbal stabilizer, a quadruped robot, and an autonomous vehicle in performing distinguishable motion for the dataset construction.
Various structured or semi-structured environments on The Hong Kong University of Science and Technology (HKUST) campus, including the lab, garden, canteen, corridor, escalator, and outdoor road, are examined in the dataset.
Also, the collected sequences present several environmental changes caused by external light, moving objects, and scene texture. These issues are challenging to SLAM algorithms.

Third, besides ground-truth poses, we also provide ground-truth maps of most indoor sequences. 
We consider that measuring the mapping accuracy is crucial for evaluation.
We also benchmark several state-of-the-art (SOTA) SLAM systems, including two vision-based methods and four LiDAR-based approaches.
To benefit the community, the dataset will be publicly released: \url{https://ram-lab.com/file/site/multi-sensor-dataset}.

\section{Related Work}
\label{sec:related_work}



There are extensive datasets for robotic perception.
Here, we introduce related works with a focus on SLAM.

Several datasets were specifically designed for one type of sensor.
Mueggler \textit{et al.} \cite{mueggler2017event} proposd the event camera dataset for the purpose of overcoming illumination and motion blur issues caused by frame cameras.
Pomerleau \textit{et al.} \cite{pomerleau2012challenging} proposed the point cloud dataset that covers a large spectrum of environmental structures to challenge registration algorithms.
Handa \textit{et al.} \cite{handa2014benchmark} promoted the research on RGB-D cameras by publishing the ICL-NUIM dataset. 

Complementing vision sensors with inertial measurements, visual-inertial odometry (VIO) approaches can tremendously improve camera tracking accuracy and robustness.
Relevant datasets have been reported.
Burri \textit{et al.} \cite{burri2016euroc} presented the EuRoc dataset collected by a micro aerial vehicle (MAV) in an industrial environment and a room.
Schubert \textit{et al.} \cite{schubert2018tum} put forward the TUM VI benchmark by collecting handheld sequences with a careful photometric calibration forwards.

The DARPA challenge has driven the development of autonomous vehicles. 
Huang \textit{et al.} \cite{huang2010high} presented the MIT DARPA dataset with over $90km$ sequence. 
Geiger \textit{et al.} \cite{geiger2013vision} presented the KITTI driving benchmark where diverse perception tasks are explored.
There are other datasets targeting at long-term navigation \cite{barnes2020oxford} and urban challenges \cite{wen2020urbanloco}.

Several datasets were collected by handheld devices and other types of ground robots.
Ramezani \textit{et al.} \cite{ramezani2020newer} collected the Newer College Dataset with a handheld device.
The NCLT dataset \cite{carlevaris2016university} facilitated the long-term SLAM research by collecting sequences in a college campus, over $147.4km$ traverse and $15$ months. 
The M2DGR dataset covers various challenging scenarios such as entering lifts and indoor-outdoor traverse \cite{yin2021m2dgr} with a ground robot.
Zhu \text{et al.} \cite{zhu2018multivehicle} proposed a multi-vehicle dataset for event-based perception.

Table \ref{tab:related_work_dataset} compares existing datasets with our work. 
In summary, our dataset is more complete from three aspects:
\textit{1)} raw and rich sensory measurements; 
\textit{2)} data collection on three different platforms including a legged robot;
\textit{3)} ground-truth trajectories and 3D maps for algorithm evaluation.

\section{System Overview}
\label{sec:system}



This section introduces sensors used in our dataset and how we achieve the spatio-temporal calibration between each sensor.
Fig. \ref{fig:sensor_picture} shows the handheld device equipped with multiple sensors and how it is mounted on three data collection platforms.

\subsection{Sensor Configuration}
\label{sec:system_sensor_configuration}
Sensors' characteristics can be found in Table \ref{tab:sensor_list}.
We use the Intel NUC to run sensor drivers, attach timestamps of sensor messages, and record messages into ROS bags on the Ubuntu system.
The PC uses an i$7$ processor, $1$TB solid-state drive (SSD), and $64$GB DDR4 memory.
Below, we provide detailed description of these sensors.

\subsubsection{3D LiDARs}
We configure the OS$1$-$128$ LiDAR to provide accurate measurements of surrounding environments.
This LiDAR has two attractive properties.
First, an internal synchronized IMU outputs $100$Hz linear accelerations and angular velocities.
Second, it additionally outputs depth images, signal images, and ambient images of surroundings. 


\subsubsection{Stereo Frame Cameras}
Two FILR BFS-U3-31S4C global-shutter color cameras are mounted at two sides on the system, facing directly forward. 
They are synchronized by an external trigger and capture high-resolution images at $20$ fps.
Their exposure time is set as fixed values to minimize the relative latency.
Our experiments show that the average difference in timestamps of these images is below $1ms$.

\subsubsection{Stereo Event Cameras}
Two event cameras are also configured.
They possess several desirable properties: high temporal resolution, high dynamic range, and low power consumption.
The cameras have a $346\times 260$ resolution and an internal high-rate IMU output.
Event cameras are synchronized using the trigger signal generated from the left camera (master) to deliver sync pulses to the right (slave) through an external wire.
But there is no way to synchronize the image acquisition (around $10$-$20ms$ offset). 
To suppress the LiDAR's laser light, both cameras are equipped with additional infrared filters. 
For indoor sequences, we manually set and fix the APS exposures, which helps to minimize the latency between cameras. 
For outdoor sequences, we use auto-exposure to avoid over- or under-exposure.

\begin{figure}[t]
	\centering
	\subfigure[Garden]
	{\label{fig:scene_garden}\centering\includegraphics[width=.272\linewidth]{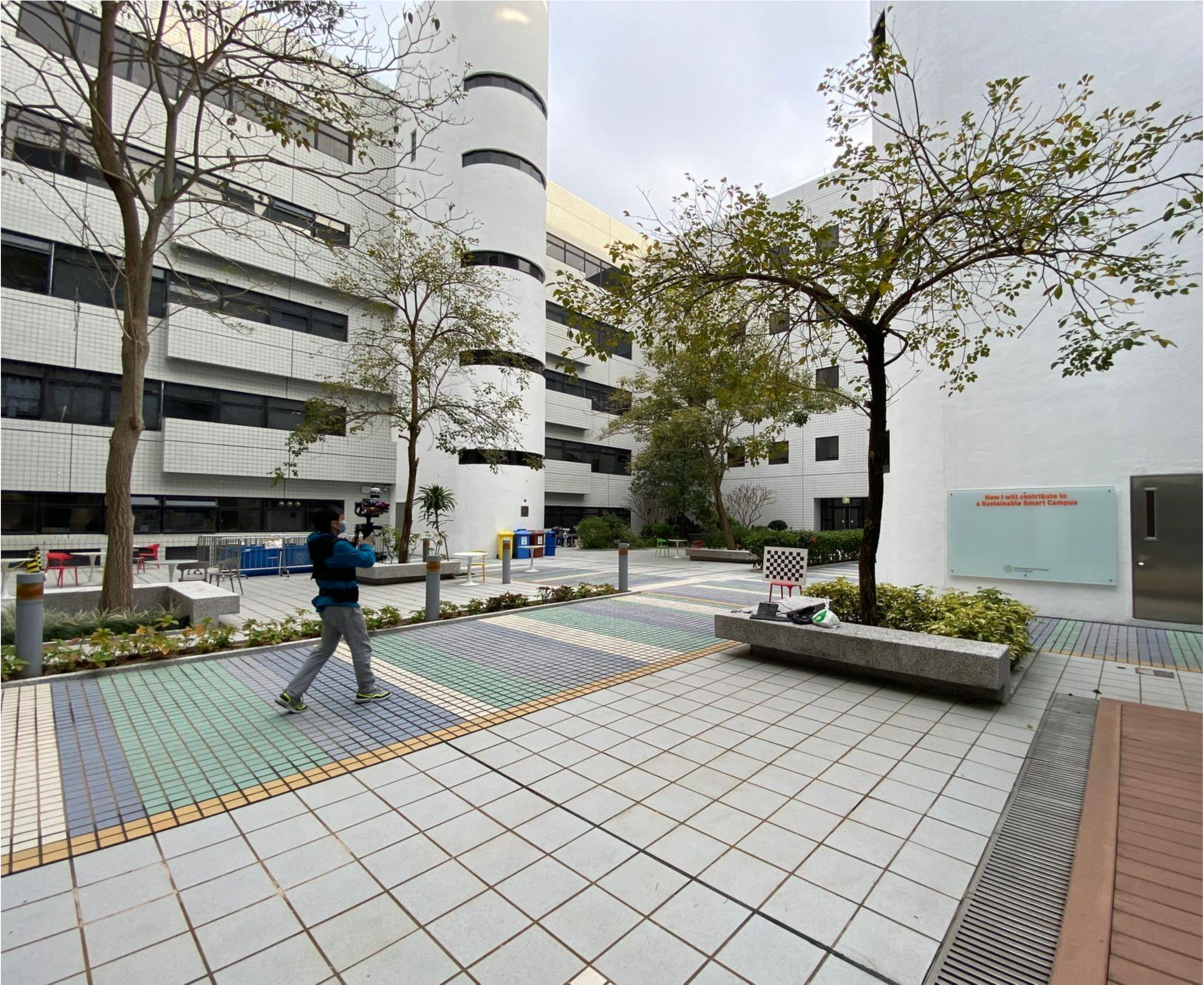}}
	\subfigure[Building]
	{\label{fig:scene_building}\centering\includegraphics[width=.288\linewidth]{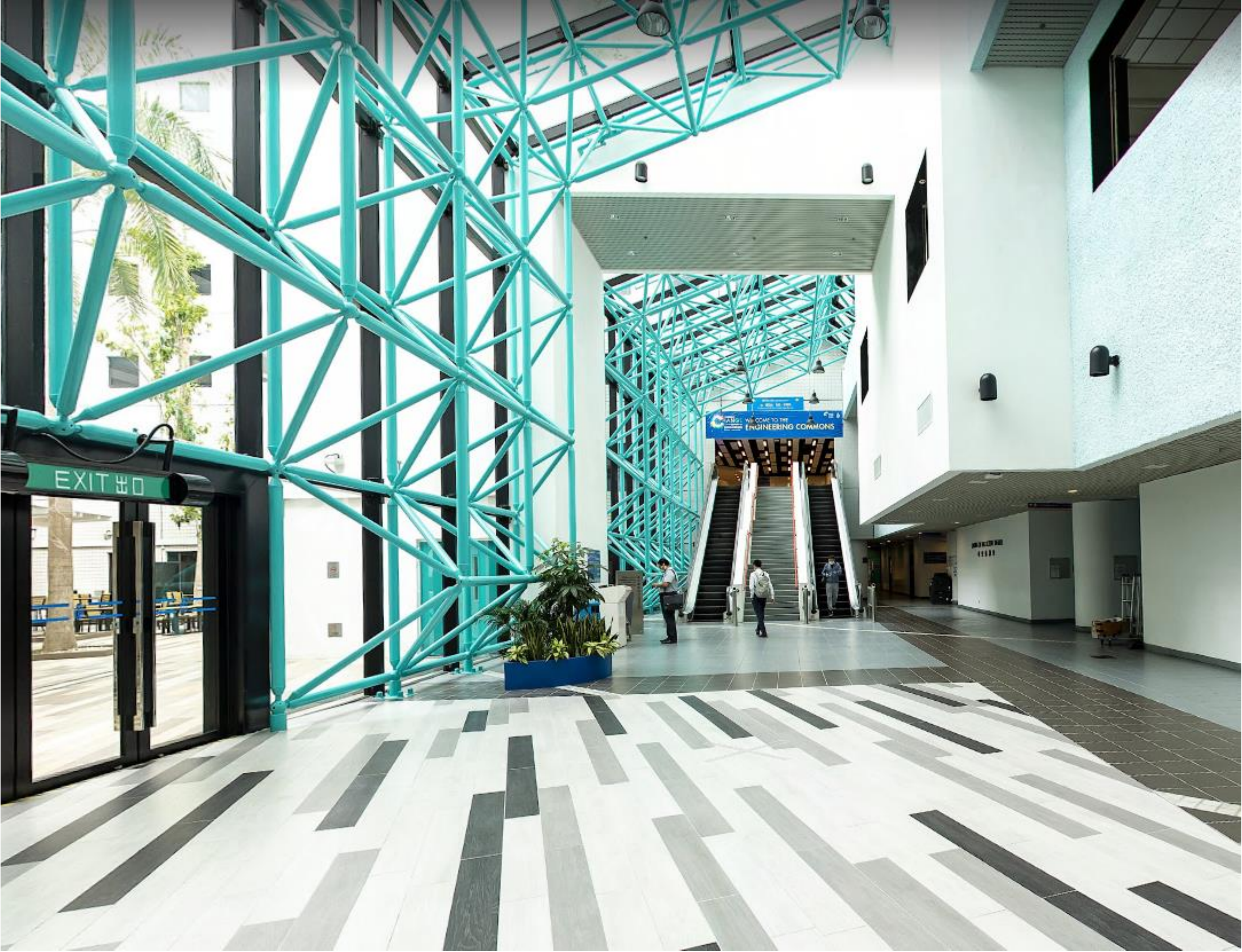}}
	\subfigure[Campus Road]
	{\label{fig:scene_campus}\centering\includegraphics[width=.392\linewidth]{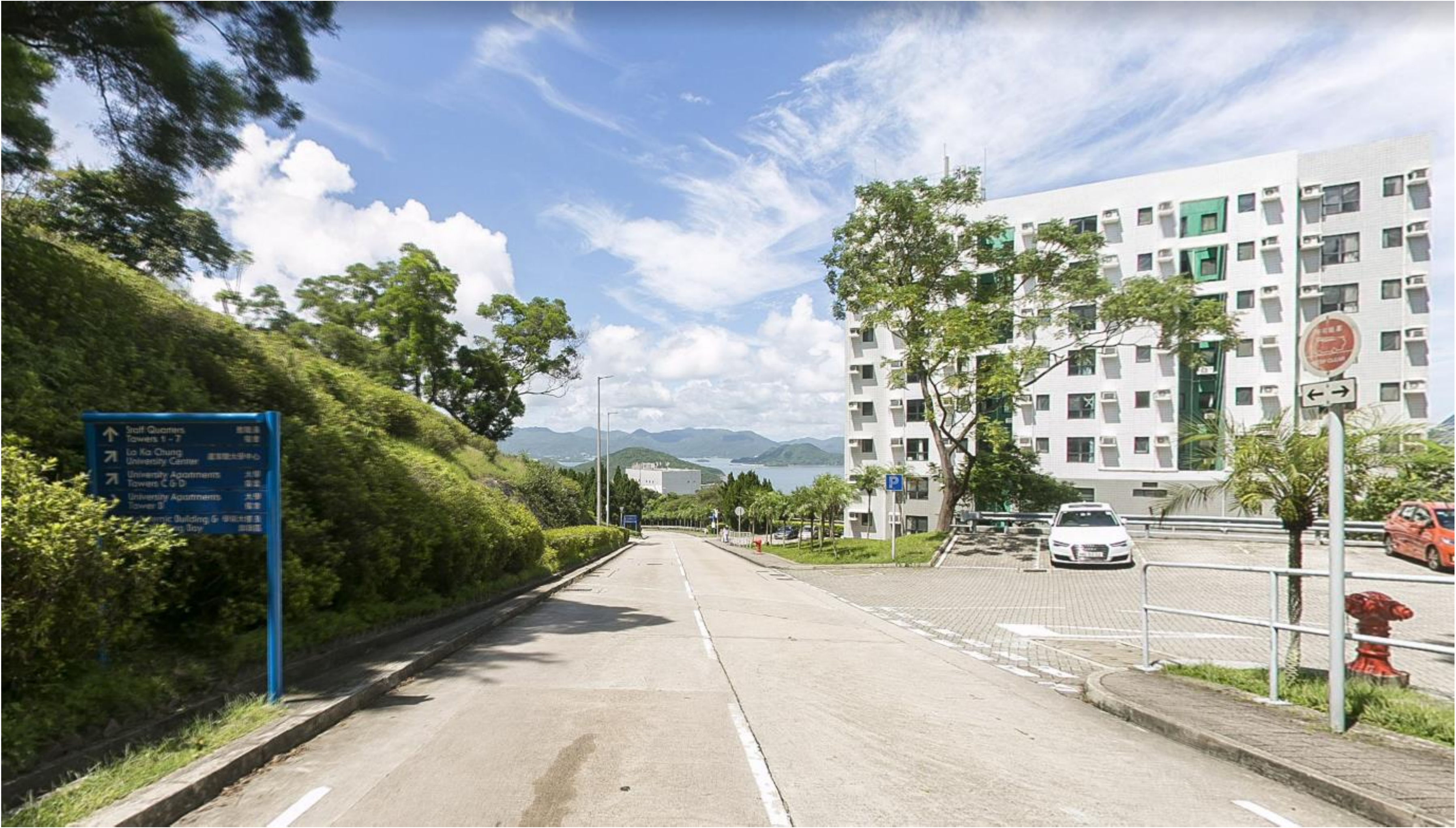}}
	\caption{Scene Images of places of several sequences.}
	\label{fig:scene_imagess}
	\vspace{-0.5cm}
\end{figure}

\subsubsection{Inertial Measurement Unit}
A tactical-grade STIM$300$ IMU that is rigidly mounted below the LiDAR is employed as the main inertial sensor of the system.
It features a high update rate ($200$Hz) and low noisy and drifting measurements.
Its bias Instability is around $0.3^{\circ}/h$.

\subsubsection{Global Positioning Systsem}
We additionally install a ZED-F9P RTK-GPS device on the top of the LiDAR.
In outdoor scenes, the GPS is activated and provides accurate latitude, longitude, and altitude readings. But it may sometimes become unstable due to buildings' occlusion.

\subsection{Sensor Calibration}
\label{sec:calib}
We carefully calibrate \textit{intrinsics} of individual sensors, \textit{extrinsics}, and overall time latency between sensors in advance.
We define the coordinate system of the STIM$300$ IMU as the \textit{body frame}.
We provide calibration data and reports in the dataset website.


\subsubsection{Clock Synchronization}
\label{sec:synchronization}
We use an FPGA to generate an external signal trigger to synchronize clocks of all sensors.
This can guarantee data collection across multiple sensors with minimum latency.
The FPGA receives a pulse-per-second (PPS) signal from the GPS and outputs $200,20,10$Hz signal to the IMU, cameras, and LiDAR, respectively.
The FPGA switches to use its internal clock to enable the time synchronization in GPS-denied scenes.



\begin{table}[]
	\centering
	\caption{Sensors and characteristics}
	\vspace{-0.2cm}
	\renewcommand\arraystretch{1.0}
	\renewcommand\tabcolsep{5pt}
	\scriptsize
	\begin{tabular}{ll}
	\toprule
	Sensor & Characteristics \\
  \midrule[0.03cm]
	
	3D LiDAR &
	\begin{tabular}[c]{@{}l@{}}
		OS$1$-$128$, $120m$ range@$\bm{10}$Hz; FOV: $45^{\circ}$vert., $360^{\circ}$horiz. \\
		Image: $1028\times 128$@$\bm{10}$Hz\\
		IMU: ICM$20948$@$\bm{100}$Hz, $9$-axis MEMS, intrinsic calibrated \\
	\end{tabular} \\	
	\midrule[0.01cm]

	\begin{tabular}[c]{@{}l@{}}
		Frame \\
		Camera
	\end{tabular} &
	\begin{tabular}[c]{@{}l@{}}
		Stereo color cameras: $2$ FILR BFS-U$3$-$31$S$4$C \\
		Resolution: $1024\times 768$, global shutter@$\bm{20}$Hz \\ 
		FOV: $66.5^{\circ}$vert., $82.9^{\circ}$horiz. \\
	\end{tabular} \\
	\midrule[0.01cm]

	\begin{tabular}[c]{@{}l@{}}
		Event \\
		Camera
	\end{tabular} &
	\begin{tabular}[c]{@{}l@{}}
		Stereo color event cameras: $2$ DAVIS346 \\
		Resolution: $346\times 240$;\ \ 
		FOV: $67^{\circ}$vert., $83^{\circ}$horiz. \\
		IMU: MPU$6150$@$\bm{1000}$Hz, $6$-axis MEMS, intrinsic calibrated \\
	\end{tabular} \\
	\midrule[0.01cm]

	IMU &
	\begin{tabular}[c]{@{}l@{}}
		STIM$300$@$\bm{200}$Hz, Bias Instability $0.3^{\circ}/h$, Allan Var. @$25^{\circ}C$ \\
	\end{tabular} \\
	\midrule[0.01cm]

	GPS &
	\begin{tabular}[c]{@{}l@{}}
		ZED-F9P RTK-GPS@$\bm{10}$Hz, $4$ concurrent GNSS, L1/L2/L5 RTK \\
	\end{tabular} \\
  \bottomrule[0.03cm]
	\end{tabular}
	\label{tab:sensor_list}
	\vspace{-0.5cm}
\end{table}

\subsubsection{Stereo Camera Calibration}
\label{sec:calib_stereo_camera}
Intrinsics and extrinsics of our stereo frame and event cameras are estimated using the Matlab calibration toolbox, where the pinhole camera and radial-tangential distortion model are used.
We move the sensor suite before a checkerboard to collect a sequence of images. 
We evenly sample images as the calibration data and manually remove outliers with high reprojection errors.



\subsubsection{Camera-IMU Extrinsic Calibration}
\label{sec:calib_camera_imu}
The intrinsics of IMUs are calibrated using the Allen derivation toolbox\footnote{\url{https://github.com/ori-drs/allan_variance_ros}} that estimates the noisy density and random walk for gyroscope and accelerometer measurements.
After that, the spatial and temporal parameters of a camera w.r.t. an IMU are obtained by the Kalibr \cite{furgale2013unified}.
Our system consists of $4$ IMUs: STIM$300$, ICM$20948$ in the LiDAR, and two MPU$6050$ in the DAVIS346 event cameras.
Thus, we calibrate the intrinsics of these IMUs, and 
estimate extrinsics of these sensor pairs:
$\langle$STIM$300$, frame cameras$\rangle$, 
$\langle$STIM$300$, event cameras$\rangle$, 
$\langle$left MPU$6050$, left DAVIS346$\rangle$, and
$\langle$right MPU$6050$, right DAVIS346$\rangle$.


\subsubsection{Camera-LiDAR Extrinsic Calibration}
\label{sec:calib_lidar_camera}
Given initial extrinsics, we further refine the camera-LiDAR extrinsics.
The checkerboard is the calibration target that provides distinctive corners and boundaries for data association. 
We extend the work proposed by Zhou \textit{et al.} \cite{zhou2018automatic} by improving feature extraction and matching step.
We instead extract the outer corners of the board from point clouds and images. 
The extrinsics are optimized by minimizing the distance of all corresponding corners.

\section{Dataset Description}
\label{sec:dataset}

\begin{figure}[t]
	\centering
	\subfigure[Canteen]
	{\label{fig:sample_canteen}\centering\includegraphics[width=0.24\linewidth]{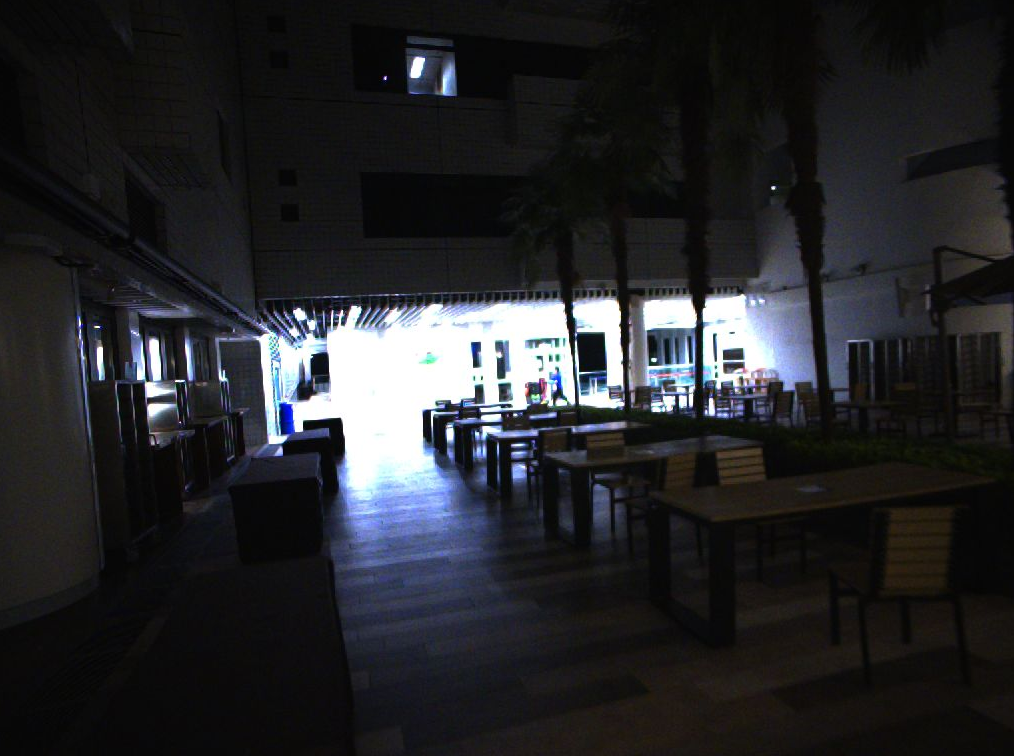}}
	\subfigure[Escalator]
	{\label{fig:sample_escalator}\centering\includegraphics[width=0.24\linewidth]{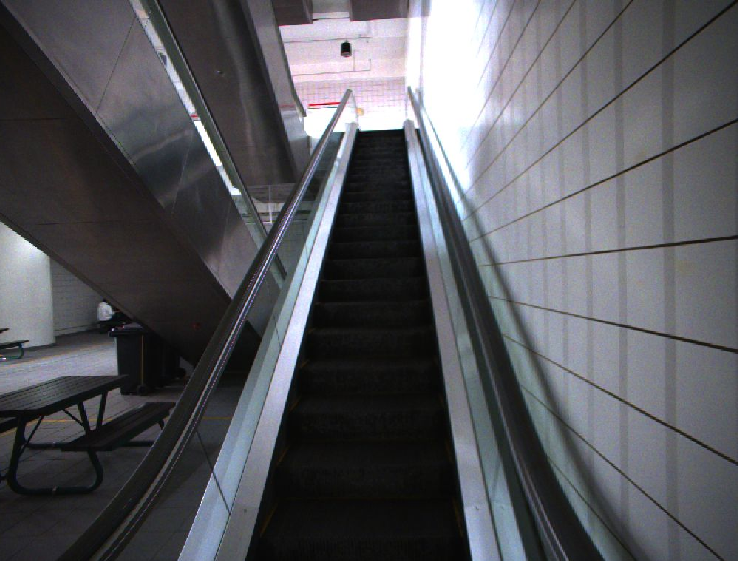}}
	\subfigure[Corridor]
	{\label{fig:sample_corridor}\centering\includegraphics[width=0.24\linewidth]{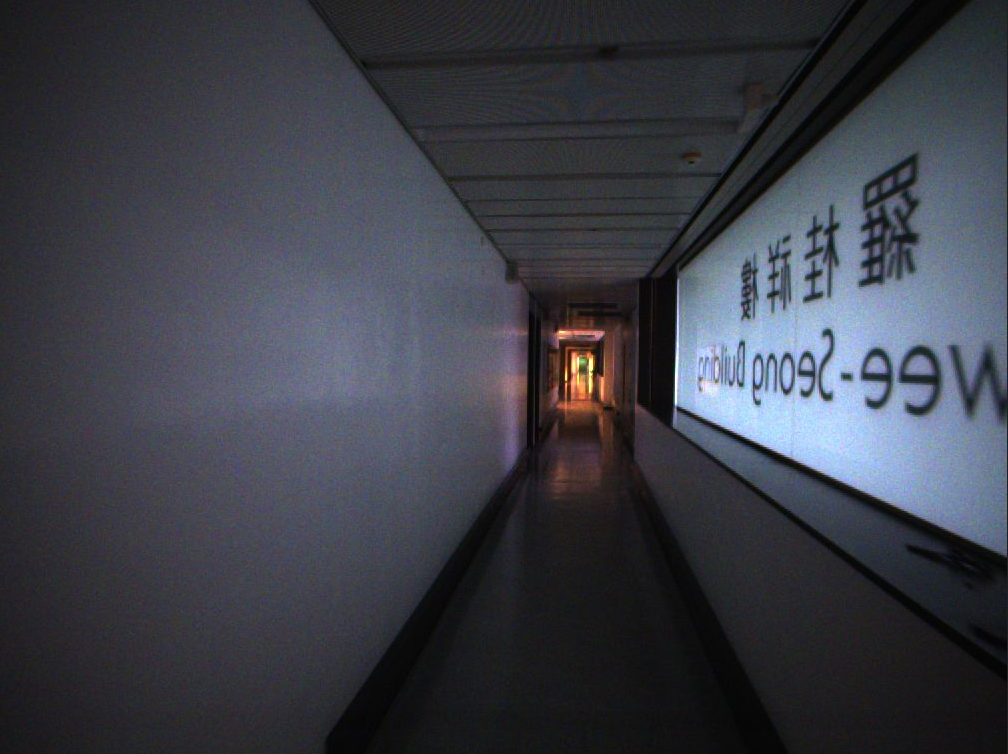}}
	\subfigure[Road]
	{\label{fig:sample_campusroad}\centering\includegraphics[width=0.24\linewidth]{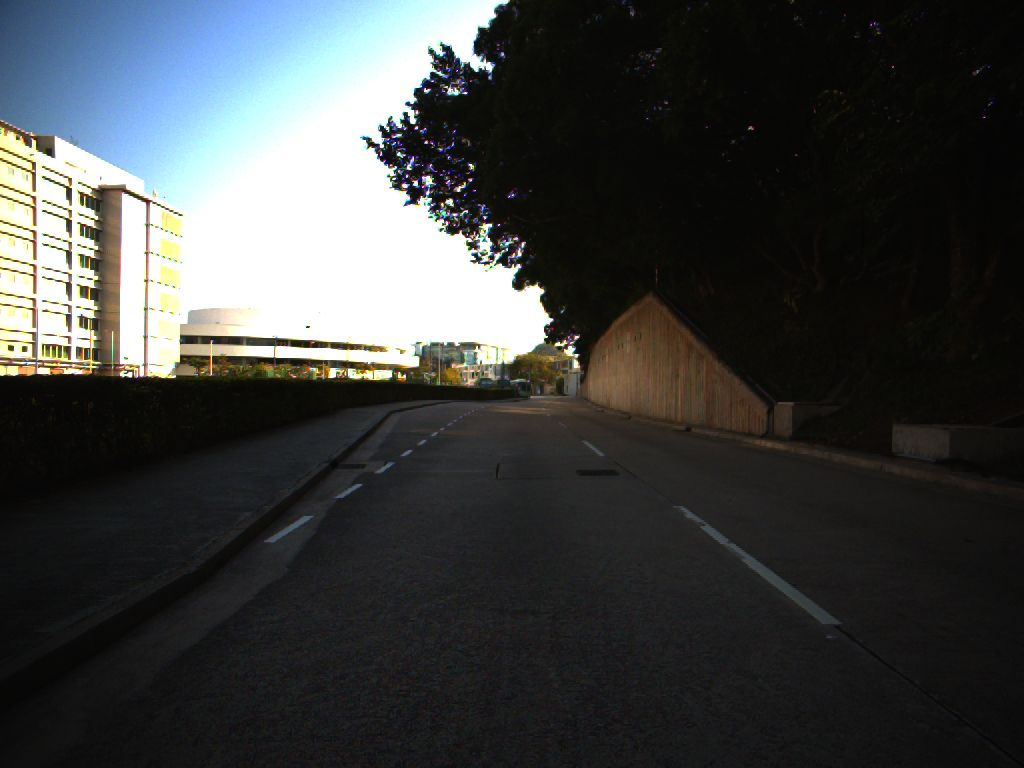}}
	\vspace{-0.1cm}
	\subfigure[MCR]
	{\label{fig:event_MCR_normal}\centering\includegraphics[width=0.32\linewidth]{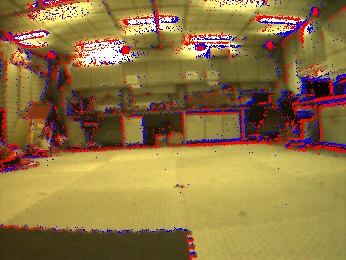}}
	\subfigure[Building]
	{\label{fig:event_building}\centering\includegraphics[width=0.32\linewidth]{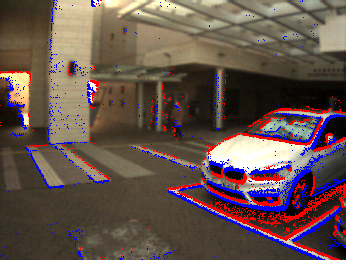}}
	\subfigure[Road]
	{\label{fig:event_campusroad}\centering\includegraphics[width=0.32\linewidth]{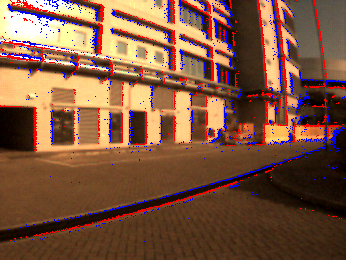}}
	\vspace{-0.1cm}
	\subfigure[Canteen]
	{\label{fig:sample_campusroad}\centering\includegraphics[width=0.435\linewidth]{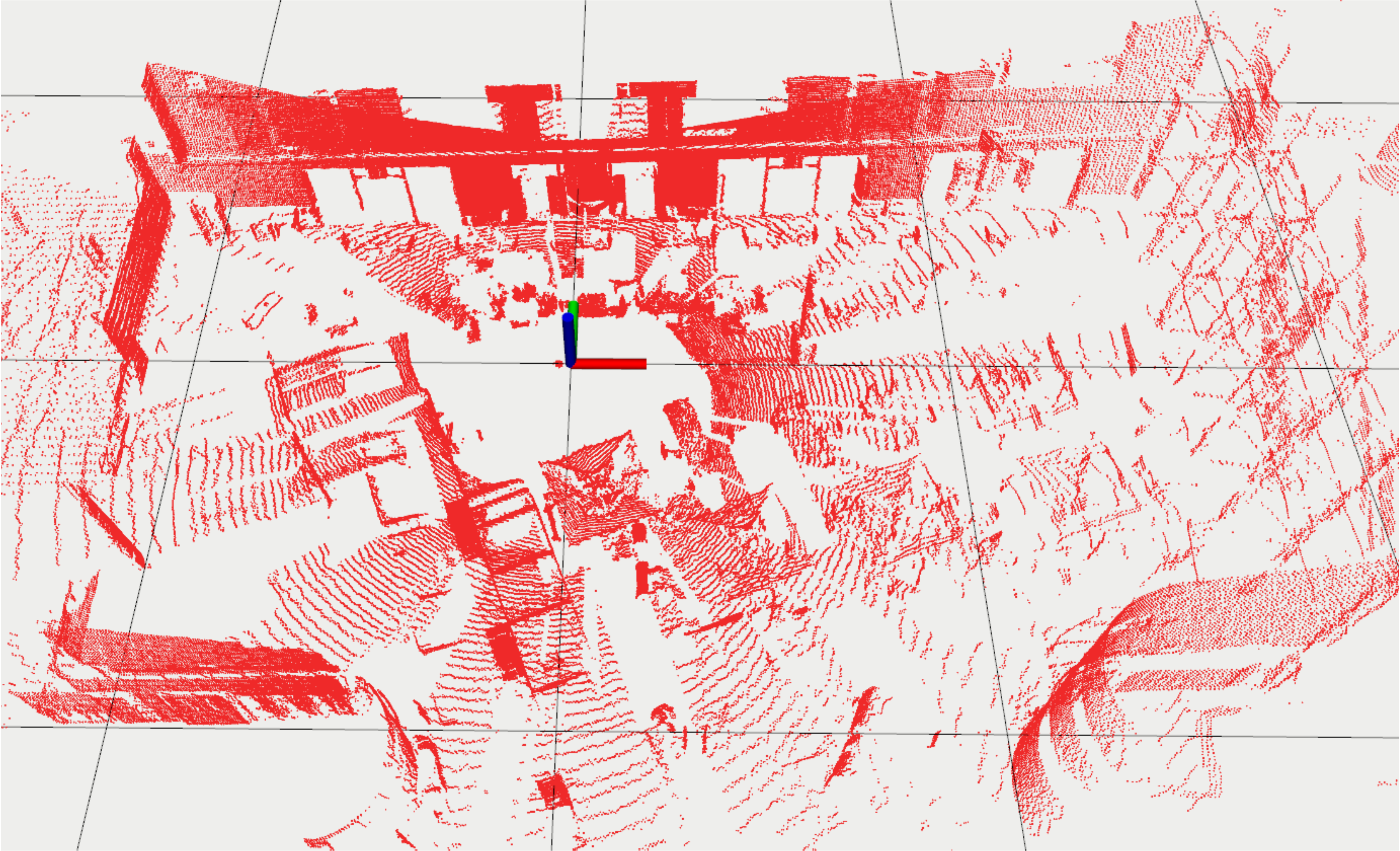}}
	\subfigure[Road]
	{\label{fig:sample_campusroad}\centering\includegraphics[width=0.55\linewidth]{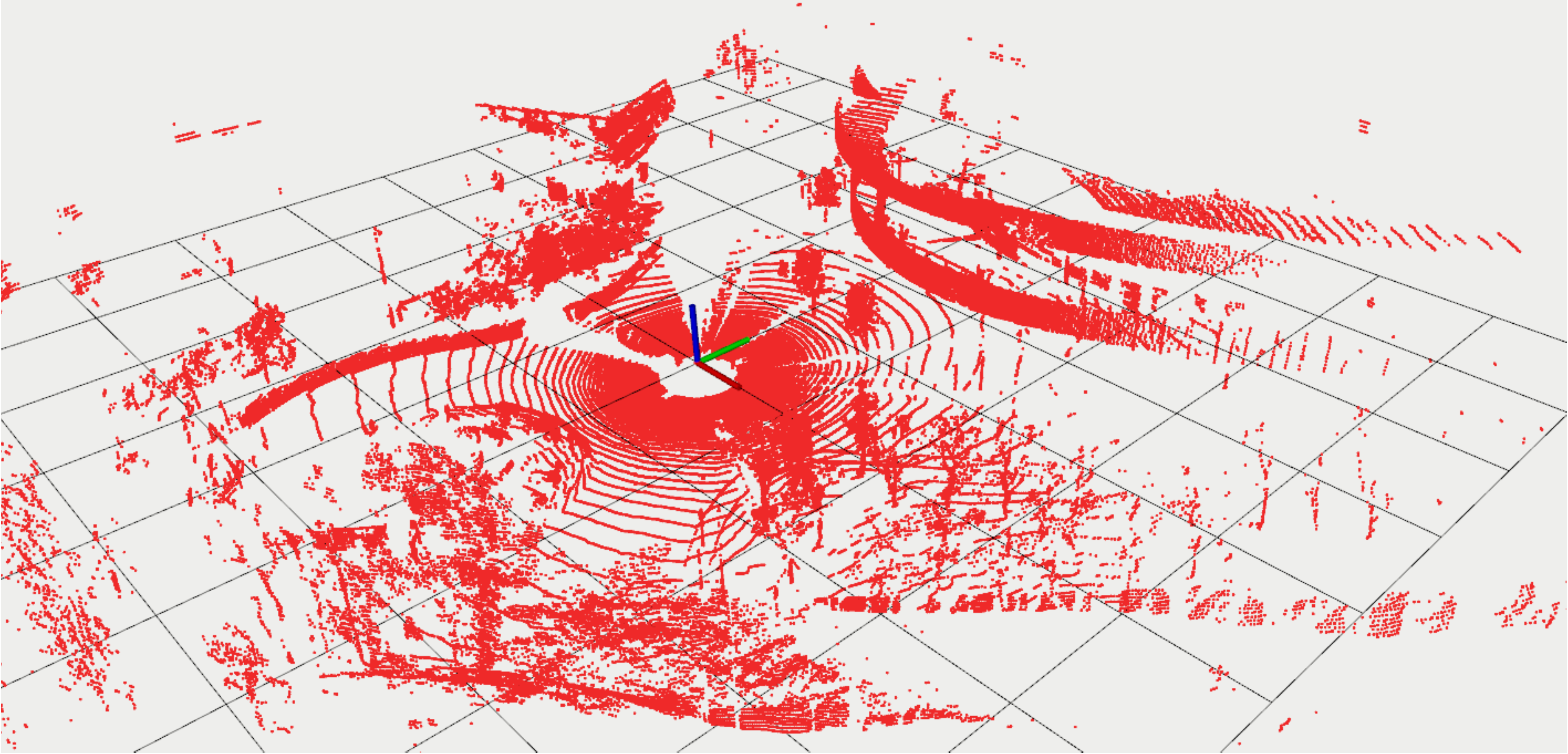}}
	\caption{Sample sensor measurements.
	(a)-(d): images captured by the frame camera.
	(e)-(f): images augmented by positive events (red) and negative events (blue).
	(h)-(i): 3D point clouds of the LiDAR.
	The grid size is $10m$.}	
	\label{fig:sample_data}
	\vspace{-0.5cm}
\end{figure}  



This section first introduces the overall features of different sequences, which stand as our basic criteria for data collection.
Details are then described, including the ground truth estimation method and dataset format.

\subsection{Sequences}
\label{sec:dataset_sequence}
The collected sequences should cover various environments, lighting conditions, motion patterns, dynamic objects, etc.
We categorize major characteristics of our collected sequences as follows:

\begin{enumerate}
  \item \textbf{Location:} Environmental locations are divided into indoors and outdoors. 
  GPS signal is available but sometimes unstable in outdoor environments.
  \item \textbf{Structure:} Structured environments can mainly be explained using geometric primitives (e.g., offices or buildings), while semi-structured environments have both geometric and complex elements like trees and sundries. Scenarios like narrow corridors are structured but may cause state estimators.
  \item \textbf{Lighting Condition:} Frame cameras are sensitive to external lighting conditions.
  Both weak and strong light may raise challenges to visual processing algorithms.
  \item \textbf{Appearance:} Texture-rich scenes facilitate visual algorithms to extract stable features (e.g., points and lines), while textureless may negatively affect the performance. Also, many events are triggered in texture-rich scenes.
  \item \textbf{Motion Pattern:} Slow, normal, and fast motion may be performed. Regarding mounted platforms, 
  the handheld device performs arbitrary 6-DoF and jerky motions, 
  the device installed on a gimbal stabilizer conducts 6-DoF but stable motions,
  the quadruped robot mostly performs planar but jerky motions. In contrast, the vehicle performs planar movements at a constant speed.
  \item \textbf{Object Motion:} In dynamic environments, several elements are moving while the data are captured. The more time of the data capture, the more deformed the elements will be (e.g., pedestrians or cars) \cite{pomerleau2012challenging}. In contrast, moving objects are few in static environments.
\end{enumerate}


\begin{table*}[]
	\centering
	\caption{Some statistics and features of each sequence}
  \vspace{-0.2cm}
	\renewcommand\arraystretch{1.1}
	\renewcommand\tabcolsep{3pt}
	\scriptsize
	\begin{tabular}{crccccccccccc}
	\toprule[0.03cm]
	Platform & Sequence & T$[s]$ & D$[m]$ & 
  $||\overline{\bm{v}}||[m/s]$ 
  & Location & Structure & Lighting & Texture & Motion & Object &
  GT Pose & GT Map\\
  \midrule[0.03cm]

  \multirow{10}{*}{Handheld} 
  & canteen\_night & $290$ & $270$ & $0.93$ & indoors & structured & weak & rich & 6-DoF & static & 6-DoF NDT & Yes \\
  & canteen\_day & $230$ & $250$ & $1.09$ & indoors & structured & normal & rich & 6-DoF & static & 6-DoF NDT & Yes \\
  & garden\_night & $280$ & $265$ & $0.94$ & indoors & structured & weak & rich & 6-DoF & static & 6-DoF NDT & Yes \\
  & garden\_day & $170$ & $173$ & $1.02$ & indoors & structured & normal & rich & 6-DoF & static & 6-DoF NDT & Yes \\
  & corridor\_day & $572$ & $669$ & $1.17$ & indoors & structured & weak & less & 6-DoF & static & 6-DoF NDT & Yes \\
  & escalator\_day & $315$ & $263$ & $0.84$ & indoors & structured & strong & rich & 6-DoF, height changes & dynamic & 6-DoF NDT & Yes \\
  & building\_day & $599$ & $666$ & $1.11$ & indoors & structured & normal & rich & 6-DoF & dynamic & 6-DoF NDT & Yes \\
  & MCR\_slow & $48$ & $50$ & $1.03$ & indoors & semi-structured & normal & rich & 6-DoF, jerky & static & OptiTrack & Yes \\
  & MCR\_normal & $45$ & $52$ & $1.26$ & indoors & semi-structured & normal & rich & 6-DoF, jerky & static & OptiTrack & Yes \\
  & MCR\_fast & $34$ & $59$ & $1.76$ & indoors & semi-structured & normal & rich & 6-DoF, jerky & static & OptiTrack & Yes \\
  \midrule[0.03cm]
  \multirow{6}{*}{
    \begin{tabular}[c]{@{}c@{}}
      Quadruped \\
      Robot
    \end{tabular}
  }
  & MCR\_slow\_$00$ & $147$ & $26$ & $0.18$ & indoors & semi-structured & normal & rich & planar, jerky & static & OptiTrack & Yes \\
  & MCR\_slow\_$01$ & $127$ & $28$ & $0.28$ & indoors & semi-structured & normal & rich & planar, jerky & static & OptiTrack & Yes \\
  & MCR\_normal\_$00$ & $103$ & $48$ & $0.54$ & indoors & semi-structured & normal & rich & planar, jerky & static & OptiTrack & Yes \\
  & MCR\_normal\_$01$ & $95$ & $43$ & $0.52$ & indoors & semi-structured & normal & rich & planar, jerky & static & OptiTrack & Yes \\
  & MCR\_fast\_$00$ & $99$ & $48$ & $0.56$ & indoors & semi-structured & normal & rich & planar, jerky & static & OptiTrack & Yes \\
  & MCR\_fast\_$01$ & $121$ & $90$ & $0.83$ & indoors & semi-structured & normal & rich & planar, jerky & static & OptiTrack & Yes \\
  \midrule[0.03cm]

  \multirow{1}{*}{Apollo}
  & campus\_road & $1186$ & $1887$ & $1.62$ & outdoors & semi-structured & normal & rich & planar & dynamic & SLAM & No\\
  \bottomrule[0.03cm]
  \multicolumn{13}{l}{
    T: Total time. D: Total distance traveled.
    MCR: motion capture room.
    $||\overline{\bm{v}}||$: Mean linear velocity.
  }\\
	\end{tabular}
	\label{tab:dataset_summary}
  \vspace{-0.6cm}
\end{table*}

\begin{figure}[t]
	\centering
	\subfigure[Motion Capture Room]
	{\label{fig:scene_mcr}\centering\includegraphics[width={.496\linewidth}]{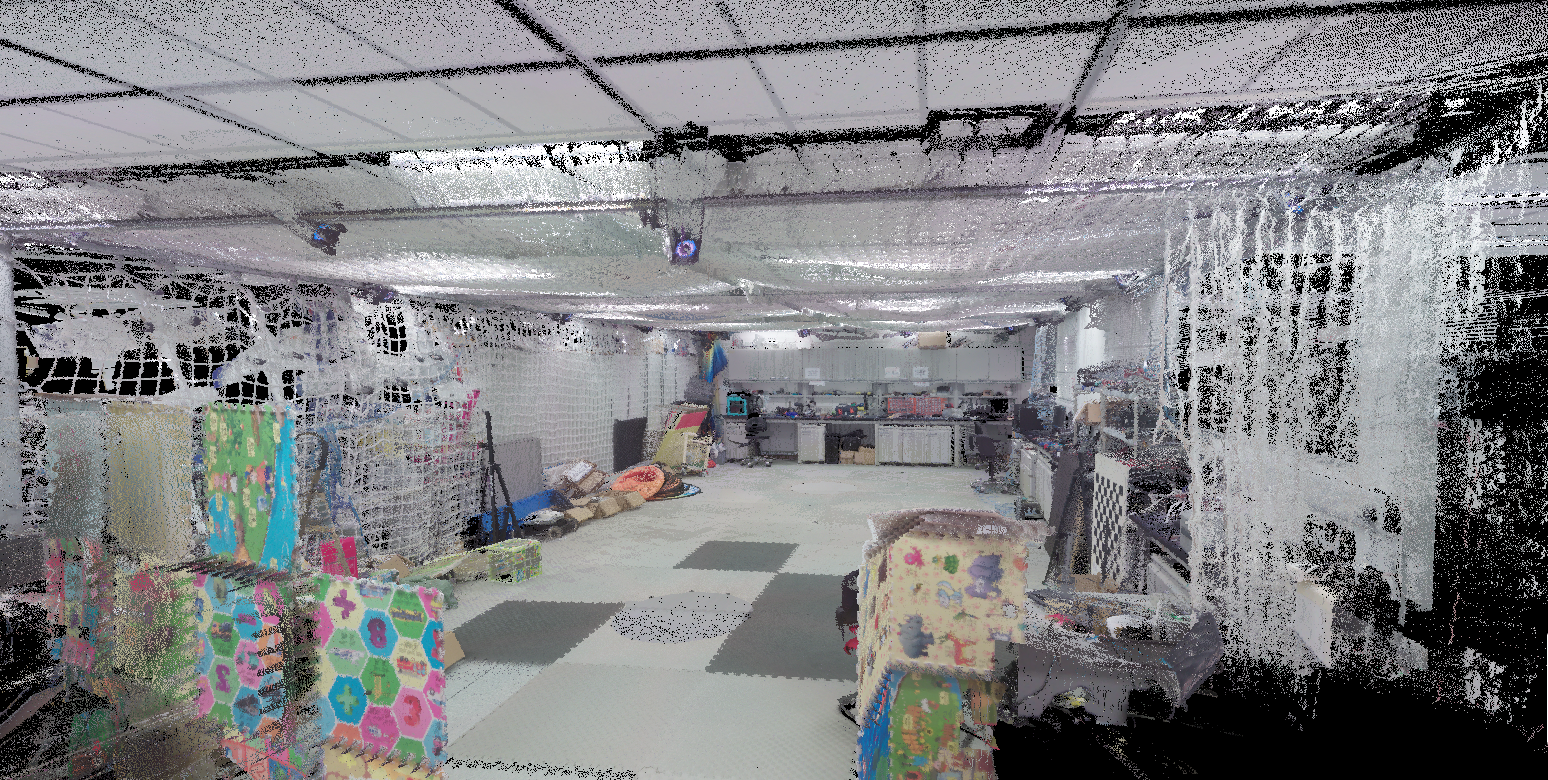}}
	\subfigure[Building]
	{\label{fig:scene_canteen}\centering\includegraphics[width={.47275\linewidth}]{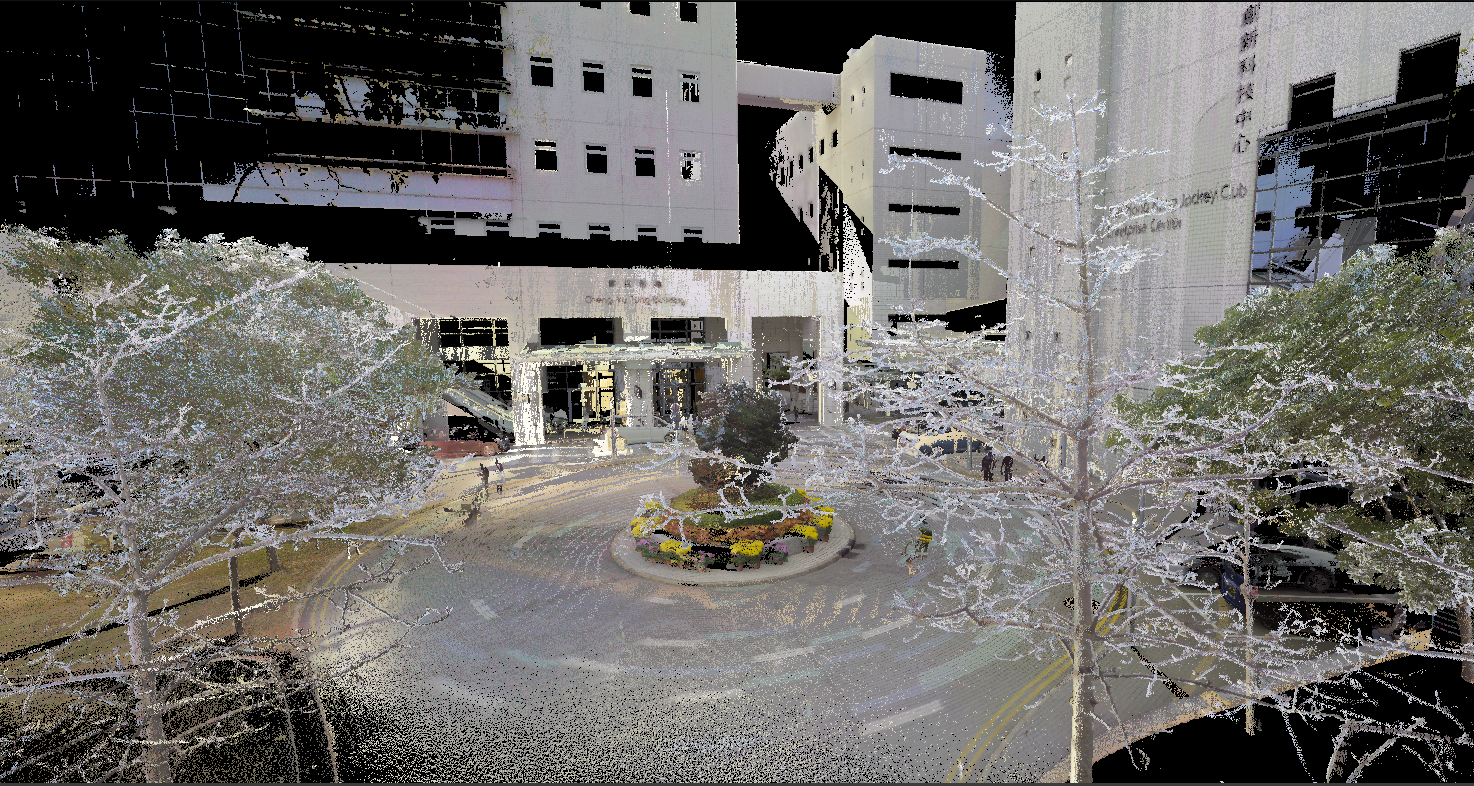}}
	\caption{Ground-truth point cloud in color of the motion capture room, corridor, and building scenario.
	Point cloud data was recorded by the Leica BLK$360$ laser scanner. They are used to generate trajectory groundtruth and evaluate algorithms' reconstruction accuracy.}
	\label{fig:gt_map}
	\vspace{-0.5cm}
\end{figure}

Table \ref{tab:dataset_summary} summaries key features of each sequence, 
Fig. \ref{fig:scene_imagess} shows several scene pictures, and 
Fig. \ref{fig:sample_data} illustrates sample sensor data.
The motion capture room is abbreviated as the MCR in the following sensors.

\subsection{Groundtruth Generation}
\label{sec:dataset_groundtruth}
Most sequences provide ground-truth poses for algorithm evaluation.
In several indoor scenes, we also provide ground-truth maps of surrounding environments.
The ground truth generation is detailed as follows:
\begin{itemize}
  \item Ground-truth maps: 
  In small- or middle-scale environments, we use the Leica BLK360 laser scanner to record the structure's high-resolution colorized 3D dense map with millimeter accuracy from multiple locations.
  Fig. \ref{fig:gt_map} visualizes three examples.  
  
  \item Ground-truth poses: 
  In the motion capture room, we use the OptiTrack system to measure the pose of the center of reflective balls at $120$Hz with millimeter accuracy. 
  The OptiTrack is directly connected with the same PC to record poses to minimize the time latency.
  The extrinsics from the balls' center to the body frame of the sensor rig are solved by the hand-eye calibration approach. 
  In middle-scale environments that are covered by the ground-truth maps, we employ the NDT-based 6-DoF localization \cite{koide2019portable} to estimate LiDAR's poses in a prior map as the ground-truth trajectory.
  In outdoor environments, we fuse the RTK GPS signal with LiDAR-inertial measurements to obtain accuracy trajectories based on the LIO-SAM \cite{shan2020lio}.
  
\end{itemize}

\subsection{Data Format and Post-Processing}
\label{sec:dataset_format}

Data were collected in the ROS environment. 
We provide both ROS bags and individual data files for better usage:
\begin{enumerate}
  \item \texttt{env.bag} is the raw rosbag obtained from the data collection process. 
  It can be parsed using ROS tools.
  \item \texttt{env\_ref.bag} is the refined rosbag where sensor data are post-processed with below steps.
  \item \texttt{data/} stores individual sensor data from the \texttt{env.bag}. Each data has its timestamp that can be retrieved from the \texttt{timestamps.txt}.
  \item \texttt{data\_ref\_kitti/} follows the KITTI format \cite{geiger2013vision} to store sensor data from \texttt{data/}. 
\end{enumerate}


We have three steps to post-process the raw data to generate the \texttt{env\_ref.bag}:
1) caused by unperfect IMUs (like the MPU$6050$), several missing measurements are linearly interpolated;
2) poses provided by the motion capture system are transformed into the body frame with the hand-eye calibration results;
and 
3) event packages are republished at around $1000$ Hz for several event-based algorithms \cite{zhou2021event}.

Unrectified RGB images are stored.
Events are stored with timestamps, pixel locations, and polarity.
IMU measurements are also stored with timestamps, gyroscope measurements, accelerometer measurements, and covariances.
Calibration parameters are stored in \texttt{yaml} files.
\section{Experiment}


As one of the applications, we can use this dataset to benchmark SOTA SLAM systems.
Here, we evaluate several open-source systems with different sensor combinations and methodologies: 
VINS-Fusion (IMU+stereo frame cameras) \cite{qin2019general}, 
ESVO (stereo event cameras) \cite{zhou2021event},
A-LOAM (LiDAR-only) \cite{zhang2014loam},
LIO-Mapping (IMU+LiDAR) \cite{ye2019tightly},
LIO-SAM (IMU+LiDAR) \cite{shan2020lio},
and
FAST-LIO2 (IMU+LiDAR) \cite{xu2022fast}.
Their data loaders are modified to fit our dataset format and also released.
We calculate the mean absolute trajectory error (ATE) of estimated trajectories w.r.t. the ground truth.
For LiDAR-based systems, we also report the mapping accuracy on two sequences by calculating the mean point-to-point error of algorithms' maps w.r.t. the ground-truth maps.


\begin{figure*}[t]
	\centering
	\subfigure[MCR\_fast\_00]
	{\label{fig:traj_mcr_fast_00}\centering\includegraphics[width=.218\linewidth]{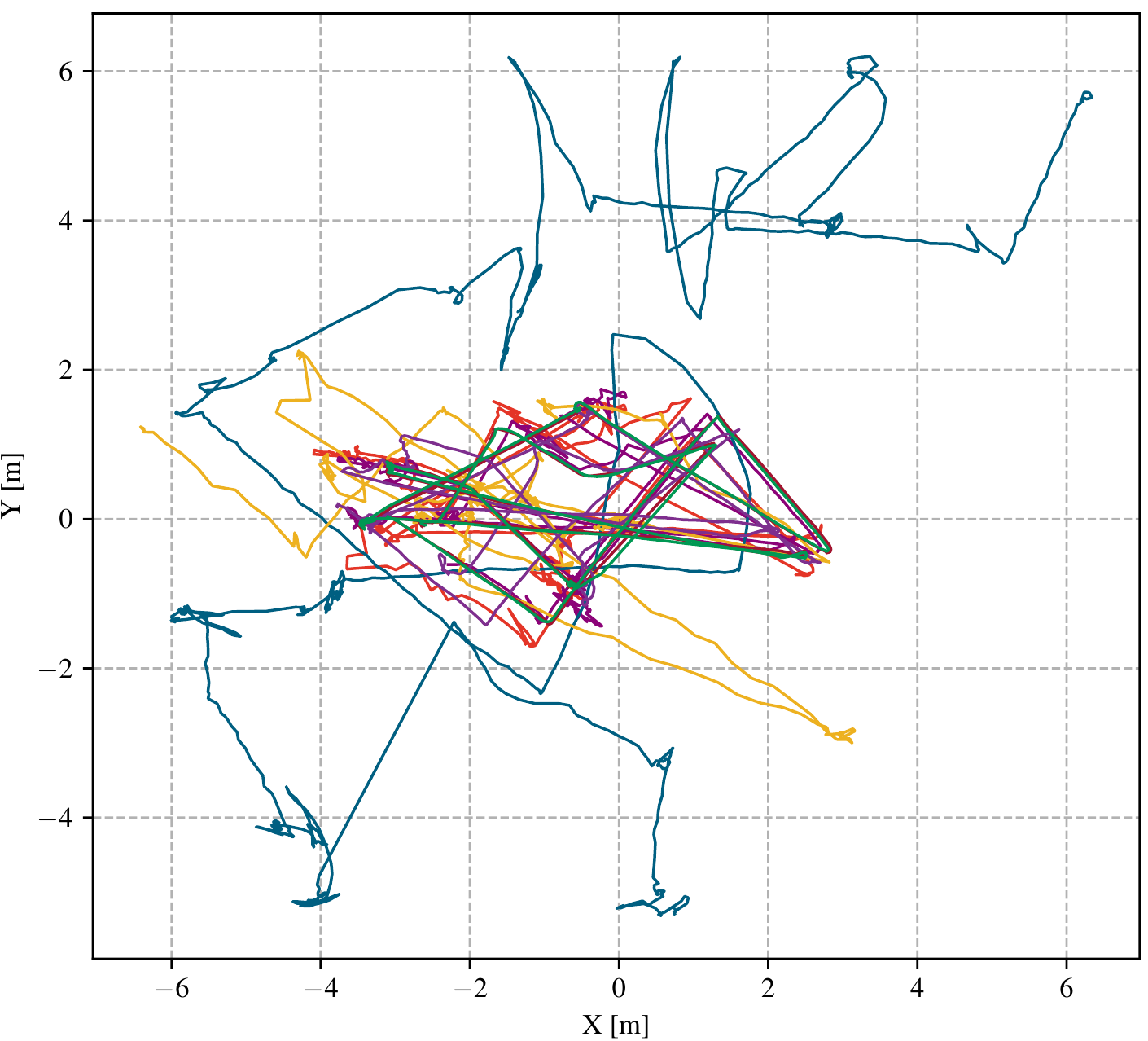}}
	\subfigure[Campus\_road\_day]
	{\label{fig:traj_campus_road_day}\centering\includegraphics[width=.14\linewidth]{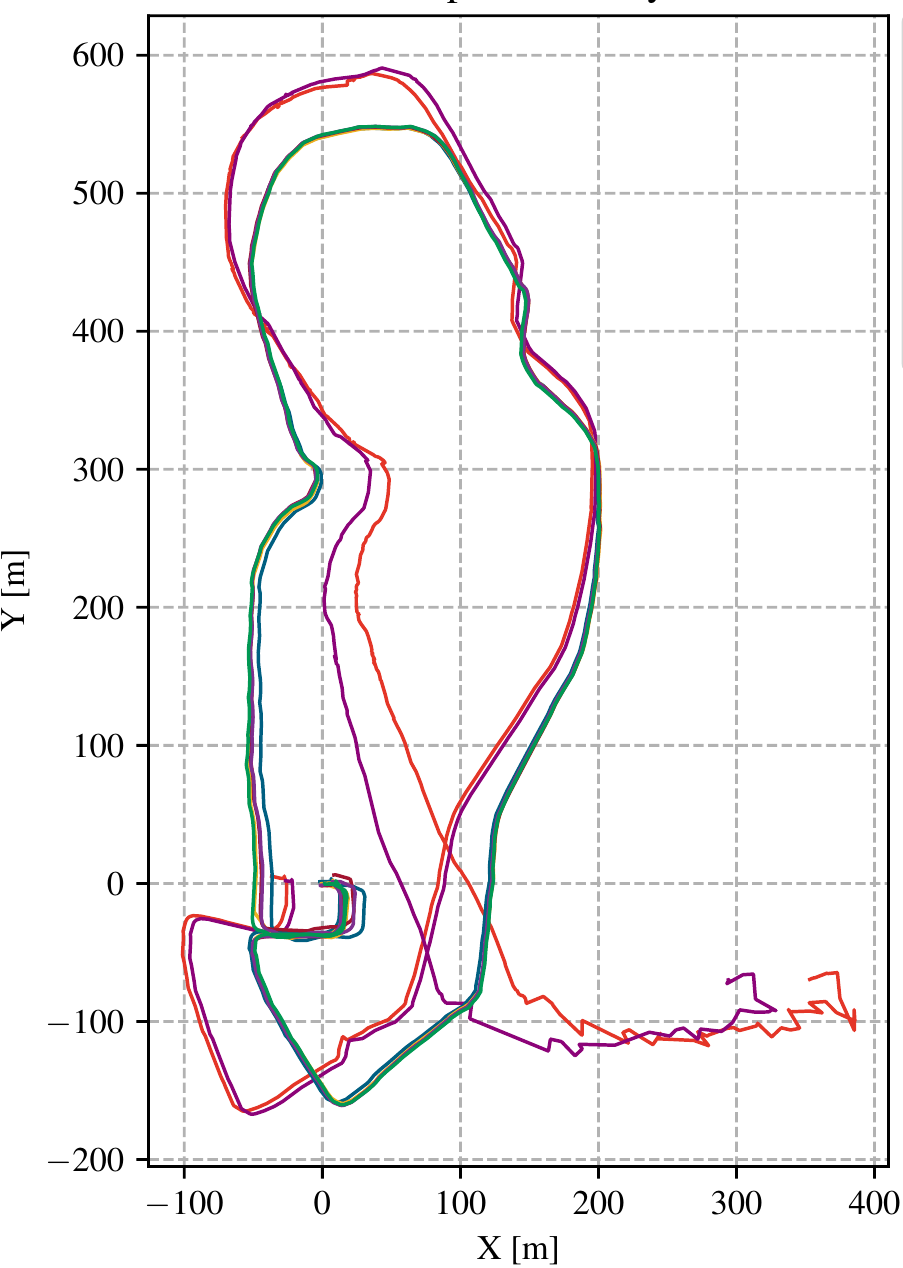}}
	\subfigure[Garden\_day]
  {\label{fig:traj_garden_day}\centering\includegraphics[width=0.2166\linewidth]{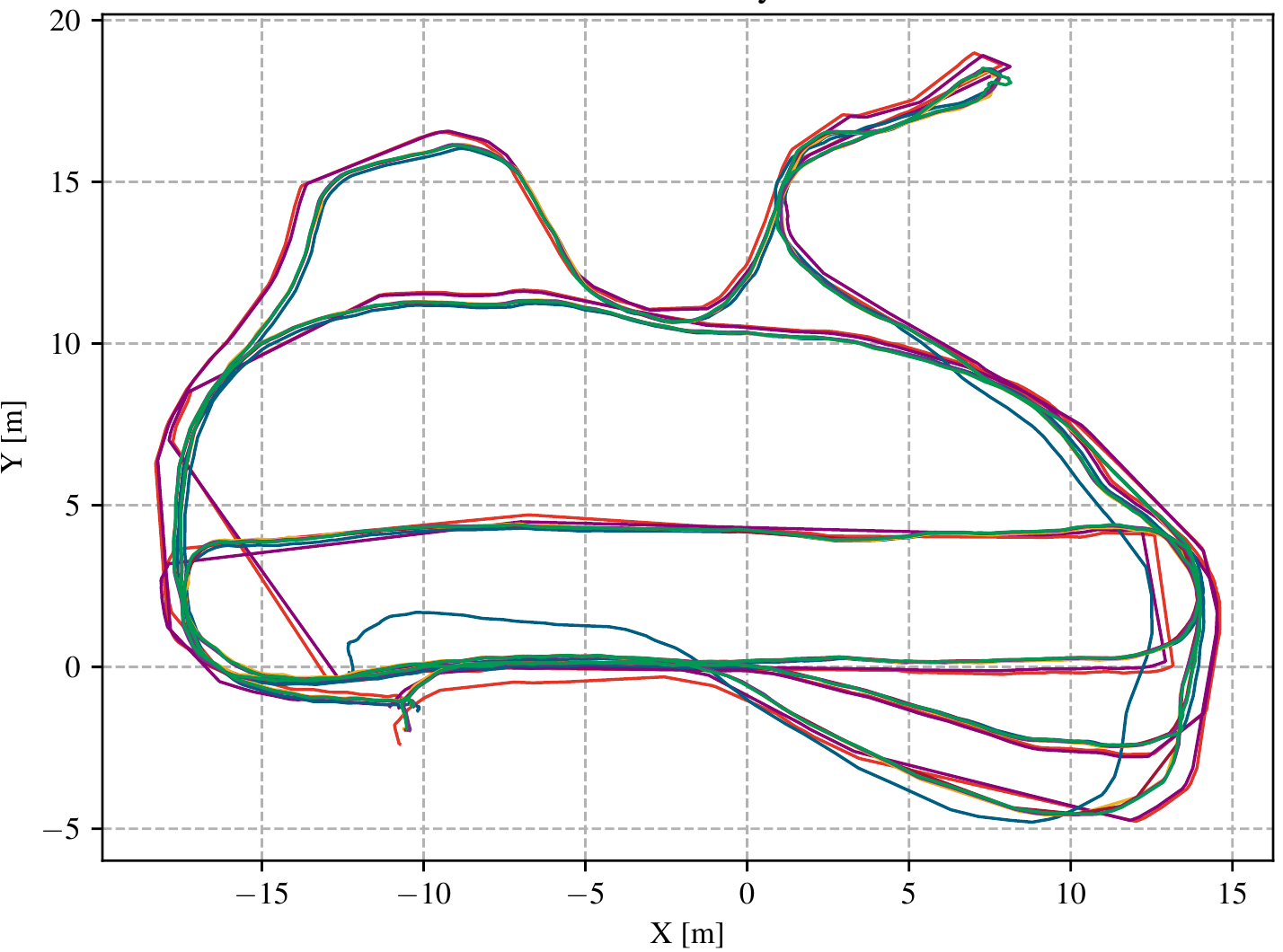}}
	\subfigure[Escalator\_day]
	{\label{fig:traj_escalator_day}\centering\includegraphics[width=.2978\linewidth]{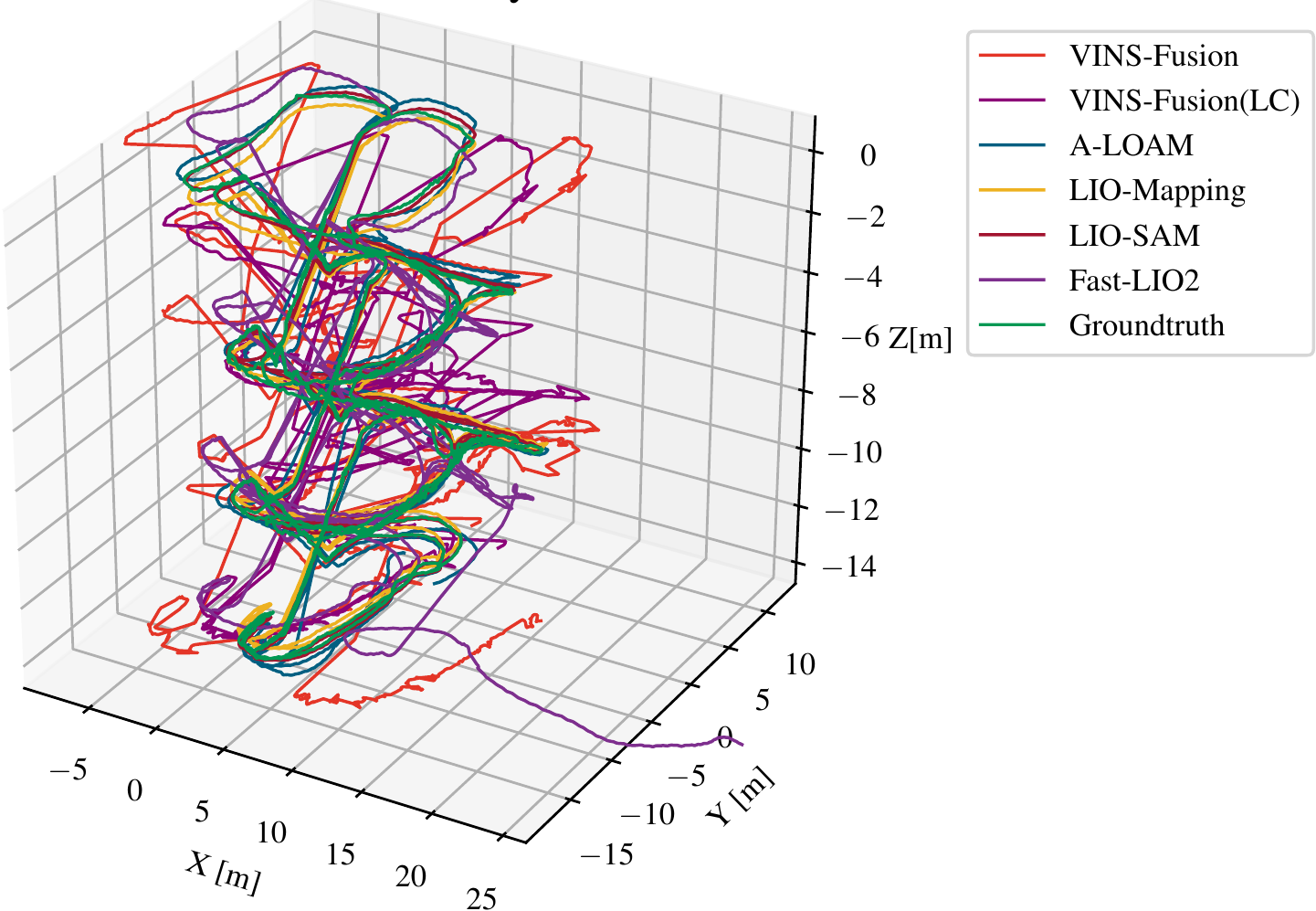}}
	\vspace{-0.2cm}
	\caption{Trajectories of the algorithms on four sequences: 
	MCR\_fast\_00, campus\_road\_day, garden\_day, and escalator\_day w.r.t. the ground truth.}
	\label{fig:exp_traj}
	\vspace{-0.4cm}
\end{figure*}

\begin{table}[t]
	\centering
	\caption{Localization accuracy.}  
  \vspace{-0.2cm}
	\renewcommand\arraystretch{1.10}
  \renewcommand\tabcolsep{2.5pt}
	\scriptsize
	\begin{tabular}{crccccc}
	\toprule[0.03cm]
  Platform &
  Sequence & 
  \begin{tabular}[c]{@{}c@{}}VINS-\\ Fusion (LC)\end{tabular} & 
  \begin{tabular}[c]{@{}c@{}}A-\\ LOAM\end{tabular} & 
  \begin{tabular}[c]{@{}c@{}}LIO-\\ Mapping\end{tabular} & 
  \begin{tabular}[c]{@{}c@{}}LIO-\\ SAM\end{tabular} & 
  \begin{tabular}[c]{@{}c@{}}FAST-\\ LIO2\end{tabular} \\

	\midrule[0.03cm]

  \multirow{9}{*}{Handheld} 
  & canteen\_night & $0.409$ & $0.067$ & $0.097$ & $\bm{0.063}$ & $0.071$ \\
  & canteen\_day & $0.691$ & $0.057$ & $0.088$ & $\bm{0.053}$ & $0.057$ \\
  & garden\_night & $0.328$ & $0.567$ & $0.242$ & $0.254$ & $\bm{0.205}$ \\
  & garden\_day & $0.518$ & $0.528$ & $0.097$ & $0.069$ & $\bm{0.068}$ \\
  & corridor\_day & $1.807$ & $\bm{0.416}$ & $1.755$ & $0.594$ & $1.563$ \\
  & escalator\_day & $2.127$ & $0.981$ & $0.346$ & $\bm{0.207}$ & $4.193$ \\
  & building\_day & $12.861$ & $1.580$ & $0.916$ & $0.222$ & $\bm{0.146}$ \\
  & MCR\_slow & $\times$ & $0.087$ & $\bm{0.042}$ & $0.063$ & $0.114$ \\
  & MCR\_normal & $0.168$ & $0.328$ & $\bm{0.052}$ & $0.082$ & $0.121$ \\
  & MCR\_fast & $\times$ & $0.416$ & $\bm{0.099}$ & $0.117$ & $\times$ \\
	\midrule[0.03cm]

  \multirow{6}{*}{
    \begin{tabular}[c]{@{}c@{}}
      Quad. \\
      Robot
    \end{tabular}
  }
  & MCR\_slow\_$00$ & $0.096$ & $0.120$ & $0.032$ & $\bm{0.023}$ & $0.047$ \\
  & MCR\_slow\_$01$ & $0.081$ & $0.054$ & $\bm{0.030}$ & $\bm{0.030}$ & $0.051$ \\
  & MCR\_normal\_$00$ & $0.094$ & $0.492$ & $0.093$ & $\bm{0.042}$ & $0.127$ \\
  & MCR\_normal\_$01$ & $0.086$ & $0.635$ & $0.390$ & $\bm{0.040}$ & $0.068$ \\
  & MCR\_fast\_$00$ & $0.264$ & $4.601$ & $2.405$ & $\bm{0.052}$ & $0.408$ \\
  & MCR\_fast\_$01$ & $0.130$ & $8.264$ & $2.210$ & $\bm{0.066}$ & $1.495$ \\
	\midrule[0.03cm]

  \multirow{1}{*}{Apollo}
  & campus\_road\_day & $77.528$ & $5.707$ & $4.122$ & $7.364$ & $\bm{4.080}$ \\
  \bottomrule[0.03cm]	
  \end{tabular}
	\label{tab:exp_localization_accuracy}
  \vspace{-0.6cm}
\end{table}


\begin{figure}[t]
	\centering
	\subfigure[Corridor\_day]
	{\label{fig:map_garden_day}\centering\includegraphics[width=.47\linewidth]{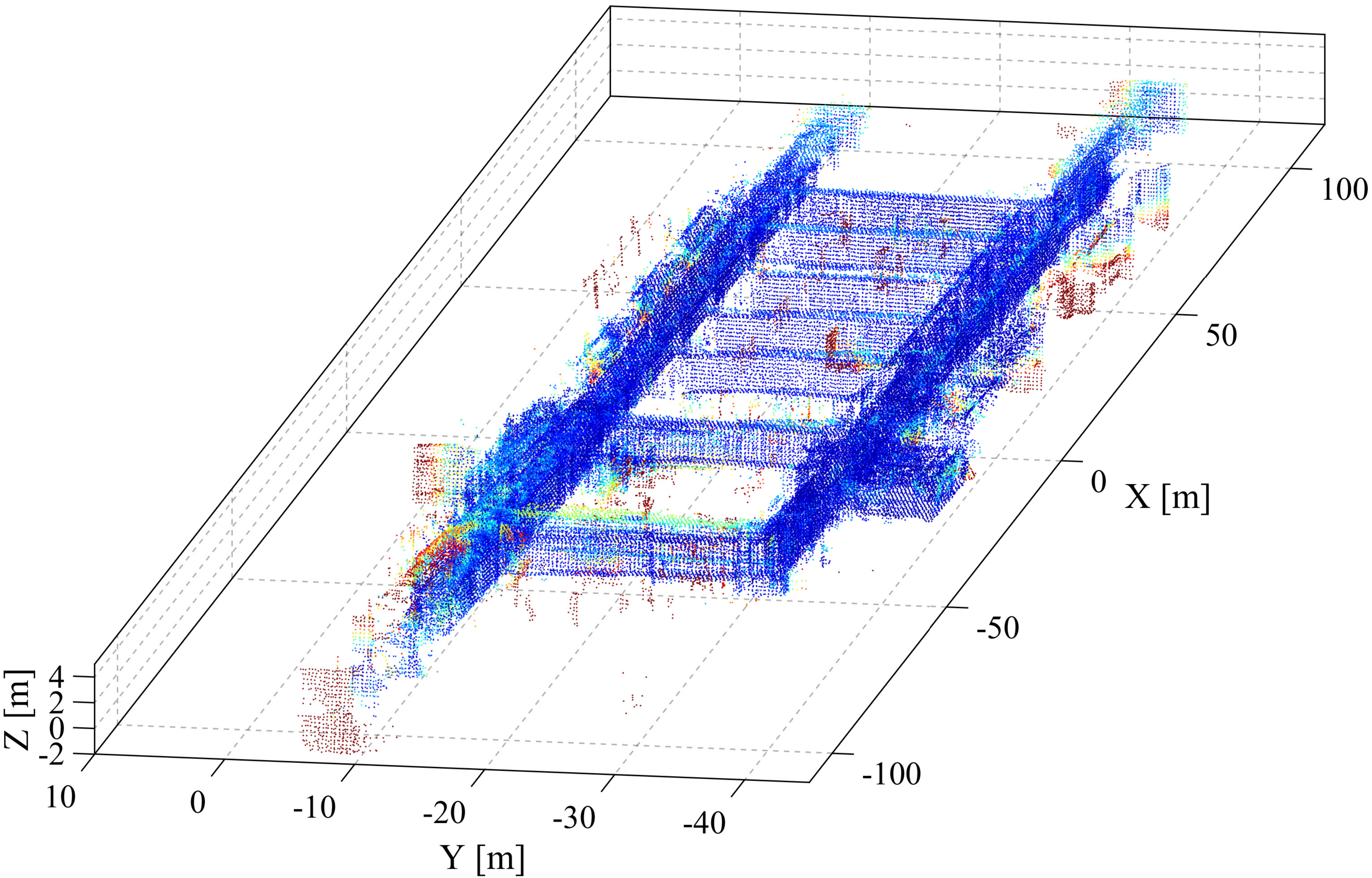}}
	\hspace{-0.3cm}
	\subfigure[Garden\_day]
	{\label{fig:map_corridor_day}\centering\includegraphics[width=.51\linewidth]{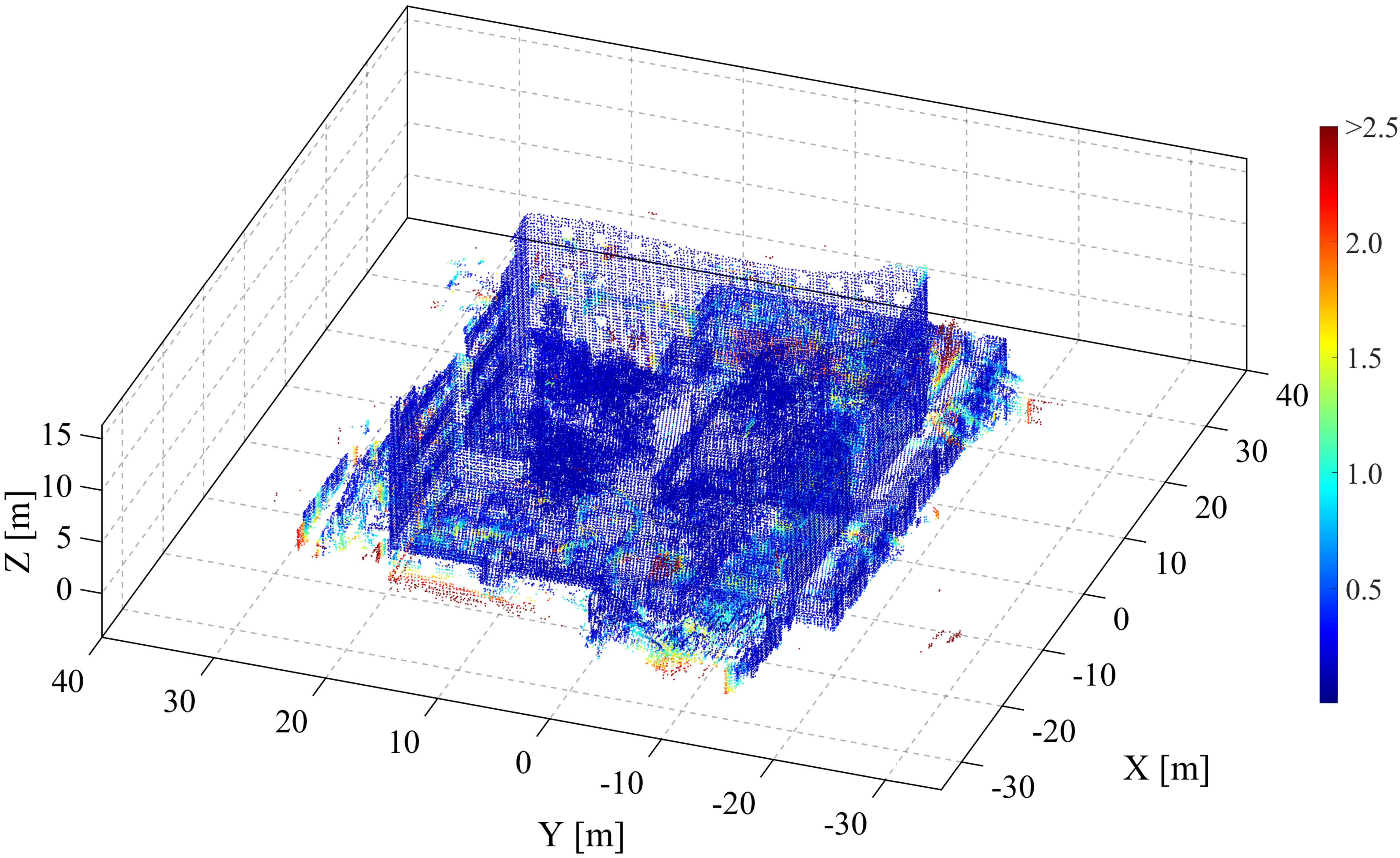}}
	\caption{Evaluation of (a) A-LOAM's and (b) LIO-SAM's mapping accuracy.}
	\label{fig:exp_mapping}
	\vspace{-0.6cm}
\end{figure}

The quantitative localization results are reported in Table \ref{tab:exp_localization_accuracy}. 
``LC'' indicates that the loop closure module is used.
``$\times$'' means that algorithms fail to finish the sequence.
ESVO's results are not shown here since it cannot finish all sequences.
It requires events to be continuously triggered to generate reliable time surface maps for camera tracking. 
But all these sequences contain textureless scenarios or static motion.
Its immediate results on mapping and tracking are shown in the dataset website.
VINS-Fusion and FAST-LIO2 fail in some cases since they cannot initialize well at the beginning of the sequence.
Without the aid of the IMU, A-LOAM cannot handle jerky and rapid motion and thus performs poorly on two MCR sequences and all sequences on the quadruped robot.
Although FAST-LIO2 has a superior real-time performance based on the filter-based state estimator and efficient tree structure, it sometimes has unreliable results on several sequences.
Surprisingly, LIO-SAM performs well on all quadruped robot-based sequences, even at large rotated and fast motion.
The corridor\_day sequence is challenging to all methods, where the scene is textureless and structureless.

We also evaluate the mapping quality of A-LOAM and LIO-SAM on the corridor\_day and garden\_day sequences. The distance map is in Fig. \ref{fig:exp_mapping}. The mean distance is $0.938m$ and $0.597m$ respectively.
Especially for the corridor mapping, A-LOAM's map has a large drift on the $z$-axis.
\section{Conclusion}
\label{sec.conclusion}



This paper presented the FusionPortable benchmark, a multi-sensor dataset from diverse campus scenes 
on various platforms. 
We advanced the self-contained and plug-and-play multi-sensor rig that significantly enhances the preception capability of mobile robots.
With the release of this dataset, we intended to challenge current SLAM approaches and encouraged future research. 
As the future work, we plan to extend this dataset beyond the campus-scale environments.
\bibliographystyle{IEEEtran}
\bibliography{ref}

\end{document}